\documentclass{article} 
\usepackage{iclr2027_conference,times}

\usepackage{amsmath,amsfonts,bm}

\def\eqref#1{equation~\ref{#1}}

\def\1{\bm{1}}

\DeclareMathAlphabet{\mathsfit}{\encodingdefault}{\sfdefault}{m}{sl}
\SetMathAlphabet{\mathsfit}{bold}{\encodingdefault}{\sfdefault}{bx}{n}

\usepackage[utf8]{inputenc}
\usepackage{amsmath}
\usepackage{amssymb}
\usepackage{amsfonts}
\usepackage{amsthm}
\usepackage{booktabs}
\usepackage{pifont}
\usepackage{textcomp}
\usepackage{multirow}
\usepackage{array}
\usepackage{colortbl}
\usepackage{xcolor}
\usepackage{mathrsfs}
\usepackage{wrapfig}
\usepackage{graphicx}
\usepackage{wrapfig}
\usepackage{subfigure}
\usepackage[colorlinks,linkcolor=black, citecolor=black]{hyperref}
\usepackage{url}

\title{DisCo: Disambiguating Multi-modal Floorplan Localization via $SE(2)$-Aware Contrastive Learning}

\author{%
\begin{minipage}{\dimexpr\textwidth-2\tabcolsep\relax}
\centering\normalfont
\textbf{Shiyong Meng\textsuperscript{1}\quad Tao Zou\textsuperscript{1}\quad Bolei Chen\textsuperscript{2}}\\[0.4em]
\textbf{Ping Zhong\textsuperscript{1}\quad Jiazhi Xia\textsuperscript{1}\quad Jianxin Wang\textsuperscript{1}}\\[0.6em]
\textsuperscript{1}Central South University\\
\textsuperscript{2}Hefei University of Technology
\end{minipage}%
}

\hypersetup{
  pdftitle={DisCo: Disambiguating Multi-modal Floorplan Localization via SE(2)-Aware Contrastive Learning},
  pdfauthor={Shiyong Meng, Tao Zou, Bolei Chen, Ping Zhong, Jiazhi Xia, Jianxin Wang}
}

\iclrfinalcopy
\begin{document}
\maketitle
\fancyhead{}
\renewcommand{\headrulewidth}{0pt}

\begin{abstract}
Visual \textbf{F}loorplan \textbf{Loc}alization (FLoc) aims to estimate camera poses by aligning visual observations with floorplans, yet it remains challenging due to structural aliasing in repetitive indoor environments. Such ambiguity causes distinct poses to exhibit similar visual-floorplan compatibility, resulting in multi-modal localization distributions with limited spatial separability and angular discriminability. Existing ray matching-based methods can generate geometry-consistent pose candidates, but lack effective mechanisms to disambiguate them. To address this issue, we propose \textbf{DisCo}, a visual-floorplan \textbf{Co}ntrastive \textbf{Dis}ambiguation framework for robust visual FLoc. First, we employ a depth-aware ray regression predictor to project monocular RGB observations into 2D ray primitives and generate multi-modal pose candidates through ray-floorplan matching. We further introduce a clustering-based hypotheses preparation strategy to extract representative FLoc hypotheses by exploiting the $SE(2)$ distribution of candidate poses. Second, we propose an $SE(2)$-aware contrastive disambiguation strategy to learn fine-grained visual-floorplan compatibility beyond geometric matching. Through structured pose perturbations, DisCo improves local pose smoothness, spatial separability, and angular discriminability by capturing implicit structural semantics in local floorplan layouts without relying on explicit semantic annotations. Experiments on three visual FLoc benchmarks show that DisCo improves pose recall over its candidate-generation baselines. On S3D (full), it achieves 75.0\% R@1 m and 74.3\% R@1 m, 30$^\circ$ without explicit semantic annotations.
\end{abstract}

\vspace{-0.5cm}
\section{Introduction}

Camera localization is a fundamental task in computer vision, supporting various applications including robotics \citep{zhong2026spatially,hunting2026learning,2026FloorPlan} and augmented reality \citep{chen2022mount}. Indoor visual localization remains challenging due to complex environments and the absence of reliable satellite signals. Conventional approaches typically rely on pre-built 3D models \citep{liu2017efficient,sarlin2019coarse,sattler2016efficient} or large-scale image retrieval databases \citep{balntas2018relocnet,2017NetVLAD}, which require substantial storage and continuous maintenance. In contrast, building floorplans provide persistent geometric references that are robust to appearance variations, making visual \textbf{F}loorplan \textbf{Loc}alization (FLoc) \citep{chen2024f3loc} a promising solution for scalable indoor localization.

\begin{figure*}[!t]
 \centering
 \includegraphics[width=0.97\linewidth]{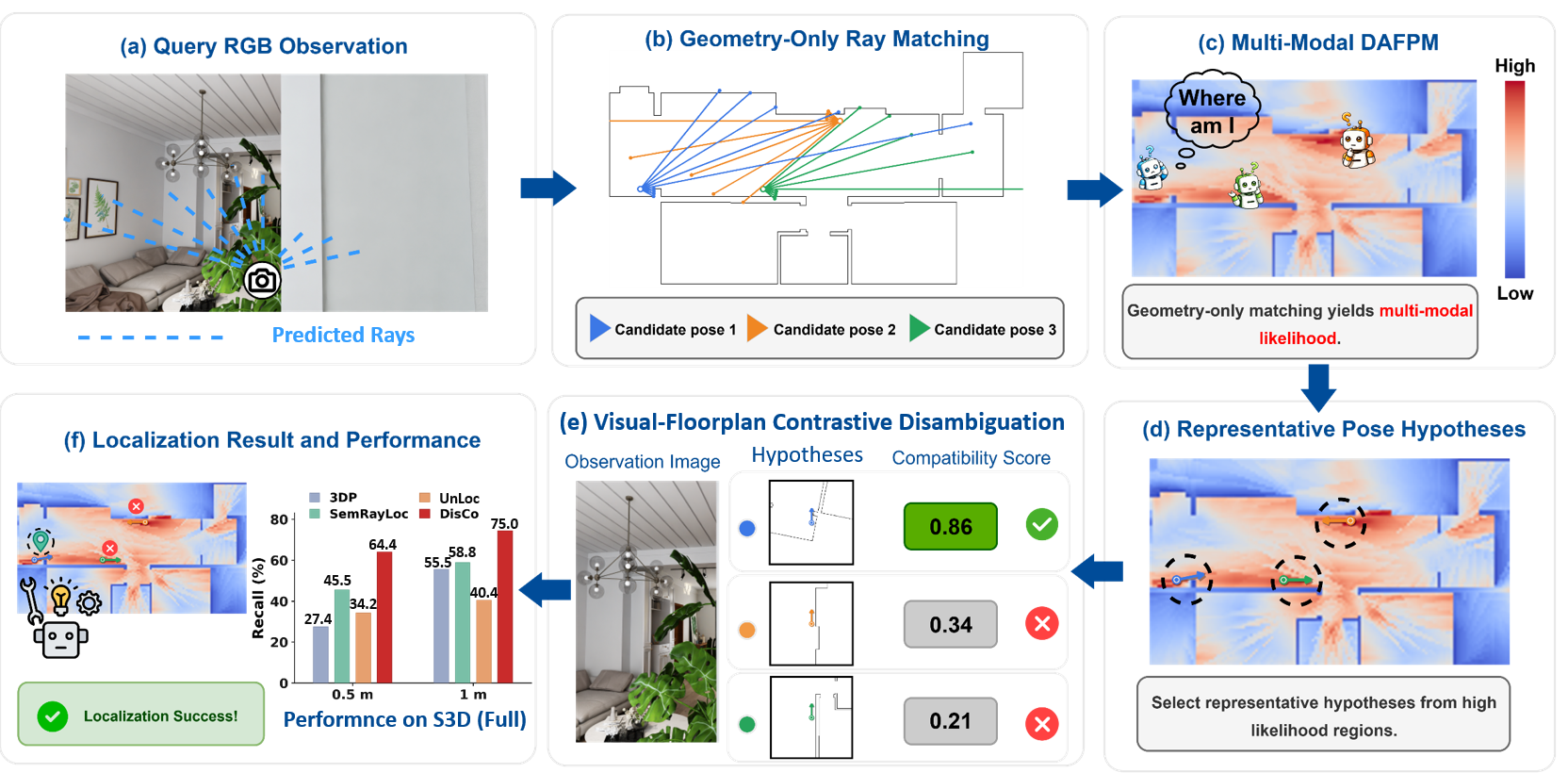}
 \vspace{-0.5cm}
 \caption{Due to modal gap and structural aliasing, ray matching-based methods generate multi-modal FLoc distributions (a)-(c). Our method first extracts representative FLoc hypotheses from DAFPM and then performs visual-floorplan contrastive disambiguation to compute compatibility scores (d)-(e), achieving superior visual FLoc without relying on explicit semantic annotations (f).}
 \label{fig1}
 \vspace{-0.7cm}
\end{figure*}

However, visual FLoc faces two fundamental challenges. The first challenge is the cross-modal gap between RGB observations and minimalist floorplans, as shown in Fig.~\ref{fig1}(a)-(b). Monocular images contain rich appearance information, such as furniture, illumination changes, and object-level clutter, whereas floorplans mainly preserve scene layouts. Existing methods bridge this gap by extracting compact geometric representations from observations \citep{min2022laser} or transforming images into ray primitives for ray-floorplan matching \citep{chen20253dp,chen2025perspective,grader2025supercharging}. Although these approaches can generate geometry-consistent pose candidates, they mainly focus on candidate retrieval, while the ambiguity among candidates remains unresolved.

The second challenge is structural aliasing caused by repetitive indoor layouts, which severely limits \textit{spatial separability} and \textit{angular discriminability}. \textit{Spatial separability} refers to distinguishing physically distant locations with similar local structures, while \textit{angular discriminability} denotes identifying the correct orientation among symmetric observations. As shown in Fig.~\ref{fig1}(c), structural aliasing leads to multiple plausible poses with similar matching scores, resulting in inherently multi-modal localization distributions. Existing approaches introduce additional semantic cues, such as object categories or room functions, to alleviate this ambiguity \citep{grader2025supercharging}. However, explicit semantics are costly to obtain, and high-level semantic categories may still fail to capture fine-grained visual-floorplan correspondences required for precise $SE(2)$ pose disambiguation.

To address these challenges, we propose \textbf{DisCo}, a visual-floorplan \textbf{Co}ntrastive \textbf{Dis}ambiguation framework by learning $SE(2)$-aware compatibility between RGB observations and floorplan structures. First, we adopt a depth-aware \textbf{R}ay \textbf{R}egression \textbf{P}redictor (RRP) as a geometry-aware candidate generator. RRP projects monocular RGB observations into compact 2D ray primitives by suppressing irrelevant vertical visual clutter. Matching the predicted rays with floorplan layouts produces a \textbf{D}epth-\textbf{A}ware \textbf{F}Loc \textbf{P}robability \textbf{M}ap (DAFPM), which provides high-recall multi-modal pose candidates. However, directly disambiguating all candidates is inefficient since high-probability candidates usually concentrate around several local modes, as shown in Fig.\ref{fig1}(c). Therefore, we further introduce a clustering-based hypotheses preparation strategy to extract representative FLoc hypotheses from pose candidates according to their $SE(2)$ spatial and angular distributions. This process removes redundant candidates while preserving distinct ambiguity modes, providing compact and informative hypotheses for subsequent disambiguation, as shown in Fig.\ref{fig1}(d).

Second, we propose an $SE(2)$-aware contrastive disambiguation strategy to refine these FLoc hypotheses. Through contrastive learning, slack positives generated by local pose perturbations encourage local \textit{pose smoothness}, while position-level and orientation-level negatives explicitly enhance \textit{spatial separability} and \textit{angular discriminability}. Instead of relying on explicit semantic annotations, DisCo learns a visual-floorplan compatibility function between RGB observations and local floorplan structures, as shown in Fig.\ref{fig1}(e). Such compatibility captures implicit structural semantics encoded in local layouts, including spatial configurations and functional characteristics, enabling fine-grained correspondence between observations and FLoc hypotheses. Normalizing the visual-floorplan compatibility scores over the retained FLoc hypotheses yields a \textbf{D}isambiguation \textbf{P}robability \textbf{M}ap (DPM), which is fused with DAFPM to obtain the final visual FLoc result, as shown in Fig.\ref{fig2}(c).

Experiments on Gibson(f/g), S3D (full), and ZInD show gains from DisCo's hypothesis reranking, with the strongest comparative results on Gibson(f/g) and at the 1 m and joint position--orientation thresholds on S3D (full). Moreover, DisCo is complementary to existing semantic-based candidate generators \citep{grader2025supercharging} and can serve as a general ambiguity resolution module. Our main contributions are summarized as follows:

\textbf{(1)} We propose DisCo, a visual-floorplan contrastive disambiguation framework for visual FLoc. Based on ray-based candidate generation, we introduce a clustering-based hypotheses preparation strategy to extract representative FLoc hypotheses and reduce redundant ambiguity modes. \textbf{(2)} We design an $SE(2)$-aware contrastive disambiguation strategy that learns fine-grained visual-floorplan compatibility without relying on explicit semantic annotations. Through pose-structured positive and negative sampling, the proposed method improves local pose smoothness, spatial separability, and angular discriminability. \textbf{(3)} Comparative and ablation studies on three visual FLoc benchmarks demonstrate the effectiveness of the proposed reranking method. The code and checkpoints will be made publicly available to facilitate research in related fields.

\vspace{-0.3cm}
\section{Related Work}
\vspace{-0.3cm}
\textbf{Visual Localization.}
Visual localization is a fundamental problem in computer vision and has been extensively studied. Image retrieval-based methods \citep{balntas2018relocnet,2017NetVLAD} estimate camera poses by searching similar images from large-scale databases. Structure-from-motion approaches \citep{panek2022meshloc,sarlin2019coarse} construct 3D scene models and establish 2D-3D correspondences through local feature matching and robust pose estimation techniques \citep{kukelova2008automatic,fischler1981random,barath2021graph}. Scene coordinate regression methods \citep{brachmann2017dsac} learn pixel-wise 3D correspondences, while pose regression methods \citep{kendall2017geometric} directly predict camera poses from images. However, these approaches typically require scene-specific 3D models or large-scale image databases, limiting scalability and long-term deployment. In contrast, visual FLoc leverages widely available floorplans as lightweight geometric references for indoor localization.

\textbf{Visual Floorplan Localization.}
Visual FLoc is closely related to LiDAR-based \textbf{M}onte \textbf{C}arlo \textbf{L}ocalization (MCL) \citep{dellaert1999monte,chu2015you,mendez2018sedar,winterhalter2015accurate}, which estimates poses by matching sensor observations with geometric maps. However, LiDAR dependence limits its deployment on common visual platforms. Early visual FLoc methods \citep{boniardi2019robot,chu2015you,howard2022lalaloc++,howard2021lalaloc,min2022laser} explore monocular or panoramic localization by matching visual cues with floorplan structures, using handcrafted priors \citep{boniardi2019robot}, learned representations \citep{min2022laser}, or synthetic views rendered from floorplans \citep{howard2022lalaloc++,howard2021lalaloc}. However, their reliance on panoramic inputs or known camera configurations limits general monocular applications.

Recent monocular visual FLoc methods \citep{karkus2018particle,min2022laser} mainly formulate localization as matching visual observations with floorplan-derived representations. Ray matching-based approaches \citep{chen2024f3loc,chen20253dp,chen2025perspective,wuest2025unloc} further represent observations as compact geometric primitives and align predicted rays with floorplan geometry. Despite their effectiveness, these methods mainly focus on retrieving geometry-consistent candidates, while structural aliasing remains unresolved. Repetitive layouts often produce multiple candidates with similar compatibility scores, resulting in multi-modal localization distributions. To alleviate this ambiguity, semantic-enhanced methods \citep{min2022laser,mendez2020sedar,grader2025supercharging} introduce cues such as room categories, doors, and windows, while \citep{chen2025perspective} explores unsupervised multi-modal alignment to reduce annotation dependency. However, these approaches mainly improve candidate generation and lack explicit modeling of fine-grained visual-floorplan compatibility for disambiguating poses within the same environment.

Different from existing approaches, DisCo is designed as a general ambiguity resolution framework. It accepts FLoc hypotheses derived from DAFPM and then learns an $SE(2)$-aware visual-floorplan compatibility function to refine these hypotheses. By exploiting structured contrastive constraints, DisCo improves spatial separability and angular discriminability without relying on explicit semantic annotations or additional 3D scene information.

\vspace{-0.3cm}
\section{Methodology}



\vspace{-0.3cm}
\subsection{Problem Formulation} \label{3.1}
This work formulates visual FLoc as estimating a camera pose in the $SE(2)$ space by aligning a monocular RGB observation with a 2D floorplan. Given a floorplan $F$ and an RGB image $\mathcal{I}\in\mathbb{R}^{h \times w \times 3}$, the objective is to estimate the absolute camera pose $S_{\mathcal{I},F}=(x,y,\theta)$, where $(x,y)$ and $\theta$ denote the spatial position and viewing orientation, respectively. Instead of deterministic regression, we model FLoc as estimating the posterior distribution:
\setlength\abovedisplayskip{0.1cm}
\setlength\belowdisplayskip{0.1cm}
\begin{flalign} \small
\begin{aligned}
\label{eq1}
\hat{S}_{\mathcal{I},F}=\arg\max_{S_i\in\mathcal{S}}p(S_i|\mathcal{I},F),
\end{aligned}
\end{flalign}
where the continuous $SE(2)$ space is discretized into candidate poses $\mathcal{S}=\{S_i\}$, forming a probability volume $P\in\mathbb{R}^{\hat{\mathcal{H}}\times\hat{\mathcal{W}}\times O}$, where each element $P(S_i)$ represents the probability $p(S_i|\mathcal{I},F)$ for a candidate pose $S_i$. Here, $\hat{\mathcal{H}}$ and $\hat{\mathcal{W}}$ denote the number of discretized cells in the $x$ and $y$ dimensions, respectively. $O$ represents the number of orientation bins. In repetitive indoor environments, structural aliasing causes multiple spatially distant or orientation-inconsistent poses to exhibit similar visual-floorplan compatibility. A reliable FLoc method requires not only generating plausible pose candidates, but also effectively disambiguating competing candidates.

\vspace{-0.3cm}
\subsection{Stage 1: RRP-based Hypotheses Preparation} \label{3.2}

Given an RGB image, existing methods predict purely geometric rays (e.g., F$^3$Loc~\citep{chen2024f3loc} and UnLoc~\citep{wuest2025unloc}) and/or additional semantic rays (e.g., SemRayLoc~\cite{grader2025supercharging}), which are matched with floorplan-derived ground-truth rays. Despite their differences, these approaches share the same paradigm: generating multi-modal pose distribution through ray-based matching.

\begin{wrapfigure}{r}{0.6\textwidth} 
 \vspace{-0.5cm}
 \centering \includegraphics[width=1.0\linewidth]{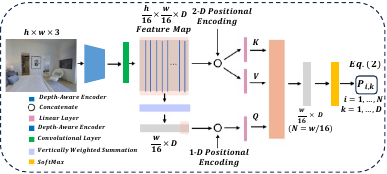}
 \vspace{-0.5cm}
 \caption{An illustration of the depth-aware RRP, which maps depth estimation expertise to probabilities across different bins in Eq. (\ref{eq2}). The RRP employs a depth-aware encoder, such as DINO-V2 \citep{oquab2023dinov2} in Depth Anything V2 \citep{2024Depth}, to extract visual features.}
 \label{fig1_2}
 \vspace{-0.3cm}
\end{wrapfigure}

\textbf{RRP-based Candidate Generator.} As shown in Fig.~\ref{fig1_2}, we adopt a depth-aware RRP~\citep{wuest2025unloc} as an example ray-based candidate generator. Specifically, the depth-aware visual features pass through a convolutional layer to generate keys and values. A vertically weighted summation is subsequently applied to form 1D queries. This operation explicitly suppresses vertical visual clutter and isolates horizontal layout structures, effectively minimizing the impact of 3D occlusions. For spatial awareness, the query's positional encodings are derived from 1D coordinates, while the positional encodings of key and value are mapped from 2D image coordinates. For each query, attention is applied across the image to predict the distance to the nearest wall. The depth prediction is formulated as a weighted sum of discrete bins:
\begin{flalign} \small
\begin{aligned}
\label{eq2}
d_i = \sum_{k=1}^{D} P_{i,k} \, d_k,\quad
d_k = \left(d_{\text{min}}^\gamma + \frac{k}{D}(d_{\text{max}}^\gamma - d_{\text{min}}^\gamma)\right)^{1/\gamma},
\end{aligned}
\end{flalign}
where $P_i \in \mathbb{R}^D$ denotes the predicted probability distribution across $D$ bins for the $i$-th ray, ensuring $\sum_{k=1}^{D} P_{i,k} = 1$ via a softmax function. The parameter $N$ ($i=1,\ldots,N$) denotes the total number of predicted rays. The terms $d_{\text{max}}$ and $d_{\text{min}}$ define the depth boundaries, while the power-law discretization parameter $\gamma$ controls the resolution distribution. The RRP is optimized using a joint $L_1$ and cosine similarity-based shape loss:
\begin{flalign} \small
\begin{aligned}
\label{eq3}
\mathcal{L}_{FLoc} = ||\mathbf{d} - \mathbf{d}^{*}||_1 + 1 - \frac{\mathbf{d}^{\top}\mathbf{d}^{*}}{\max\{||\mathbf{d}||_2||\mathbf{d}^{*}||_2,\epsilon \}},
\end{aligned}
\end{flalign}
where $\mathbf{d}$ and $\mathbf{d}^{*}$ represent the predicted and ground-truth 2D-ray depths respectively, and $\epsilon$ is a small constant to prevent division by zero.

During inference, the predicted rays are matched with floorplan-derived ground-truth rays~\citep{chen2024f3loc}, producing a probability volume $P_d\in[0,1]^{\hat{\mathcal{H}}\times\hat{\mathcal{W}}\times O}$. Instead of a deterministic prediction, $P_d$ provides a multi-modal candidate pose distribution, \emph{i.e.}, a DAFPM, as shown in Fig.~\ref{fig1}(c). In this work, we first extract Top-$K$ pose candidates from DAFPM and then propose DisCo to disambiguate them.

\textbf{Clustering-based Hypotheses Preparation.} However, since high-likelihood pose candidates frequently cluster around a few local modes, directly disambiguating all Top-$K$ candidates introduces severe computational redundancy. Therefore, we perform a $SE(2)$-aware candidate refinement to extract representative FLoc hypotheses before contrastive disambiguation. We first sort all pose candidates by their DAFPM likelihoods. Then, we iteratively retain the highest-scoring unassigned candidate and suppress redundant neighbors within its local $SE(2)$ basin:
\begin{flalign} \small
\begin{aligned}
\label{eq4}
d_{SE(2)}(S_i,S_j)
=
\sqrt{
\frac{\|\mathbf{t}_i-\mathbf{t}_j\|_2^2}{\sigma_t^2}
+
\lambda_\theta
\frac{\Delta(\theta_i,\theta_j)^2}{\sigma_\theta^2}
},
\end{aligned}
\end{flalign}
where $\mathbf{t}_i=(x_i,y_i)$ and $\Delta(\theta_i,\theta_j)=\operatorname{wrap}(\theta_i-\theta_j)$. A pose candidate $S_j$ is suppressed if $d_{SE(2)}(S_i,S_j)\le \rho$. \emph{We conduct parametric studies of $\sigma_t$, $\sigma_\theta$, $\lambda_\theta$, and $\rho$ in the Appendix \ref{C1}.} The retained set $\mathcal{R}$ subsequently consists only of spatially and angularly distinct representative pose hypotheses. We experimentally find that the number of pose hypotheses typically remains within $10^2$, which is manageable. By controlling the value of $K$ or the number of FLoc hypotheses, we can strike a balance between preserving the correct pose mode and the computational cost.

\begin{figure*}[!t]
 \centering
 \includegraphics[width=1.0\linewidth]{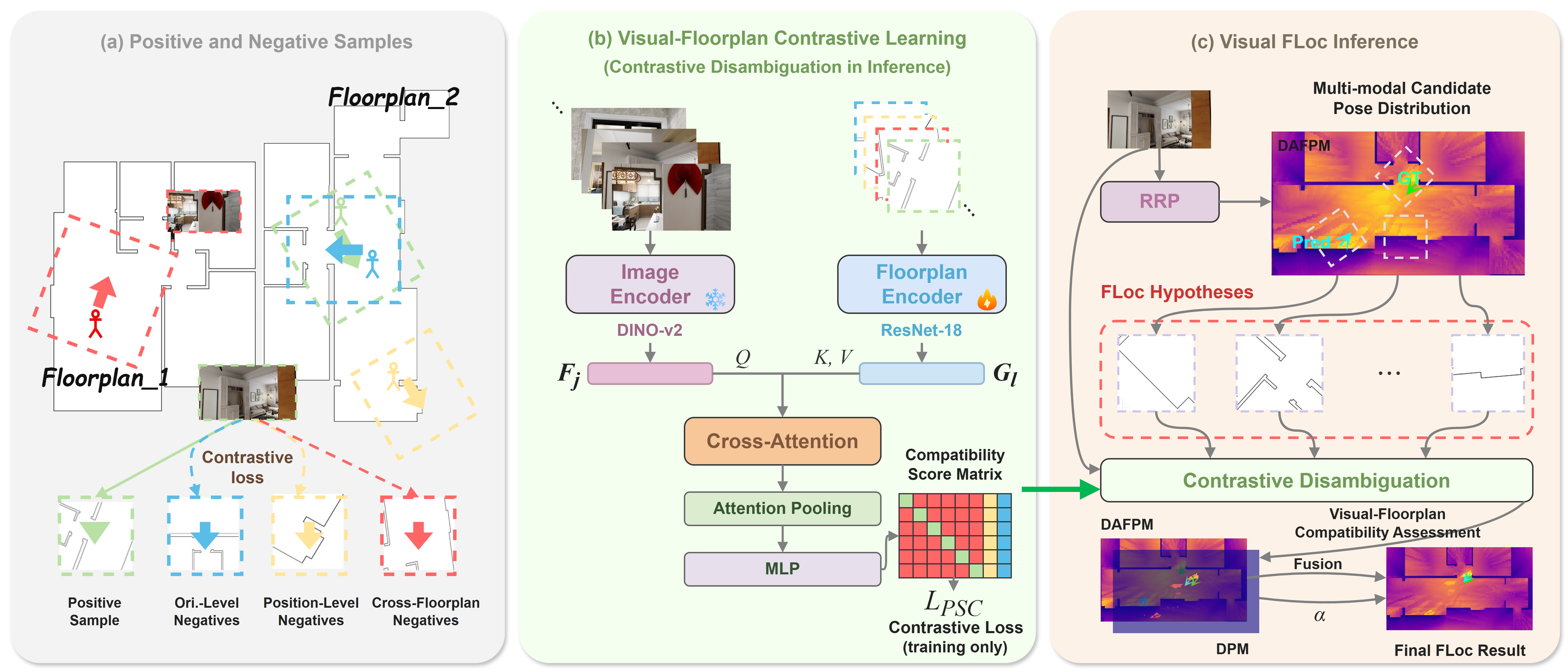}
 \vspace{-0.5cm}
 \caption{
 (a) shows the spatially perturbed sampling strategy, which constructs slack positives, position-level negatives, and orientation-level negatives according to the $SE(2)$ pose structure.
 (b) details the proposed visual-floorplan contrastive disambiguation.
 (c) DisCo performs a visual-floorplan compatibility assessment of the FLoc hypotheses derived from the DAFPM to obtain a DPM. Fusing DAFPM with DPM suppresses ambiguous hypotheses, and yields the final FLoc.
 }
 \label{fig2}
 \vspace{-0.7cm}
\end{figure*}

\vspace{-0.3cm}
\subsection{Stage 2: $SE(2)$-Aware Contrastive Disambiguation} \label{3.3}
\textbf{Visual-Floorplan Compatibility Modeling.}
For each candidate pose $S_l=(x_l,y_l,\theta_l)$, we crop the floorplan at $(x_l,y_l)$ and rotate it by $-\theta_l$, aligning the candidate heading to a canonical direction. We score the query image against this view-aligned crop to disambiguate Stage 1 hypotheses without explicit semantic annotations.

Let $\mathbf{F}_j$ and $\mathbf{G}_l$ denote image and local floorplan tokens at candidate $S_l$. Cross-attention uses $\mathbf{F}_j$ as queries and $\mathbf{G}_l$ as keys/values; residual addition and layer normalization yield candidate-conditioned tokens $\mathbf{H}_{j,l}$. Learned attention pooling and an MLP produce the compatibility score:
\begin{flalign} \small
\begin{aligned}
\label{eq5}
c(\mathcal{I}_j,F,S_l)=s_\theta(\mathbf{F}_j,\mathbf{G}_l)
=\operatorname{MLP}(\operatorname{AttnPool}(\mathbf{H}_{j,l})).
\end{aligned}
\end{flalign}
The shared network scores each pair independently (Appendix~\ref{A1}).
We construct contrastive samples according to the structures of the $SE(2)$ pose space rather than treating all poses uniformly, as shown in Fig.\ref{fig2}(a). These pose-structured constraints guide the model to learn meaningful visual-floorplan compatibility, maintaining local pose smoothness while enhancing spatial separability and angular discriminability.

\textbf{Slack Positives for Local Pose Smoothness.}
For a query image $\mathcal{I}_j$ with a ground-truth pose $S_j=(x_j,y_j,\theta_j)$, we construct a slack positive by applying a minor pose perturbation:
\begin{flalign} \small
\begin{aligned}
\label{eq6}
S_j^+ = S_j \oplus u, \quad u \sim \mathcal{U}(-B,B),
\end{aligned}
\end{flalign}
where $B$ defines the perturbation boundary for translation and orientation, and $\oplus$ denotes pose composition in $SE(2)$. The local floorplan structure cropped at $S_j^+$ is encoded into tokens $\mathbf{G}_j^+$. This slack positive establishes a local pose neighborhood around the ground truth, encouraging consistent visual-floorplan compatibility under minor pose variations.

\textbf{Position-Level Negatives for Spatial Separability.}
We construct position-level negatives $(\mathbf{F}_j,\mathbf{G}_m^{p-})$ by sampling poses with distant translations while maintaining comparable orientations. These samples often correspond to visually similar local structures at different physical positions. By explicitly distinguishing them from positives, the model learns to suppress spatially ambiguous matches and improves \textit{spatial separability}.

\textbf{Orientation-Level Negatives for Angular Discriminability.}
We construct orientation-level negatives $(\mathbf{F}_j,\mathbf{G}_m^{a-})$ by changing the viewing orientation at the correct physical position. These samples enforce the discrimination of FLoc hypotheses with correct positions but incorrect headings, thereby improving \textit{angular discriminability} and mitigating directional ambiguity.

\textbf{Cross-Floorplan Negatives.}
We retain standard cross-floorplan negatives $(\mathbf{F}_j,\mathbf{G}_m^{c-})$ by treating floorplan features from other batch samples as generic negatives. They provide coarse scene-level separation, while the proposed position-level and orientation-level negatives specifically address structural aliasing within the same floorplan.

We learn the visual-floorplan compatibility function by optimizing a single-positive \textbf{P}ose-\textbf{S}tructured \textbf{C}ontrastive (PSC) loss. For concise notation, we define the positive score and three negative partition terms as follows:
\begin{flalign} \small
\begin{aligned}
\label{eq7}
z_j^+ = \exp(s_\theta(\mathbf{F}_j,\mathbf{G}_j^+)/\tau), \quad
Z_j^c = \sum_{m \in \mathcal{N}_j^c}
\exp(s_\theta(\mathbf{F}_j,\mathbf{G}_m^{c-})/\tau),\\
Z_j^p = \sum_{m \in \mathcal{N}_j^p}
\exp(s_\theta(\mathbf{F}_j,\mathbf{G}_m^{p-})/\tau), \quad
Z_j^a = \sum_{m \in \mathcal{N}_j^a}
\exp(s_\theta(\mathbf{F}_j,\mathbf{G}_m^{a-})/\tau).
\end{aligned}
\end{flalign}
Here, $\mathcal{N}_j^c$, $\mathcal{N}_j^p$, and $\mathcal{N}_j^a$ index the three negative types. Image and floorplan tokens use a frozen DINOv2-S backbone \citep{oquab2023dinov2} and trainable ResNet-18 \citep{he2016deep}, respectively. With a learned temperature $\tau$ and batch $\mathcal{B}$, we optimize:
\begin{flalign} \small
\begin{aligned}
\label{eq8}
\mathcal{L}_{PSC}
= - \frac{1}{|\mathcal{B}|}
\sum_{j \in \mathcal{B}}
\log
\frac{z_j^+}{z_j^+ + Z_j^c + Z_j^p + Z_j^a}.
\end{aligned}
\end{flalign}
\textbf{$SE(2)$-Aware DisCo.}
At inference, the same network scores each hypothesis:
$c_m=s_\theta(\mathbf{F}_j,\mathbf{G}_m)$.
The compatibility scores are normalized to obtain the DPM:
$P_c(R_m)=\operatorname{softmax}_m(c_m)$.
We fuse the DPM with the geometric likelihood to select the pose:
\begin{flalign} \small
\begin{aligned}
\label{eq9}
\hat{S}_{\mathcal{I}_j,F}
=
\underset{R_m\in\mathcal{R}}{\arg\max}
\left(
\log P_d(R_m|\mathcal{I}_j,F)+\alpha P_c(R_m)
\right).
\end{aligned}
\end{flalign}
Here, $\alpha$ weights the normalized compatibility scores, computed without the training temperature. Image tokens are cached and candidate scoring is batched.

\vspace{-0.5cm}
\section{Experiments}
\vspace{-0.3cm}
\subsection{Experimental Setup} \label{4.1}


\textbf{Datasets.} Based on existing studies~\citep{chen2025perspective, grader2025supercharging}, we conduct comparative and ablation studies on three visual FLoc benchmarks, including the Gibson (f/g)~\citep{chen2024f3loc}, S3D (full)~\citep{zheng2020structured3d}, and ZinD~\citep{cruz2021zillow} datasets. They provide complementary evaluation settings across synthetic and real-world scenes. Gibson(g) consists of general motions, including in-place steering motions, while Gibson(f) consists of only forward motions. Therefore, Gibson(g) is intuitively more challenging than Gibson(f). S3D (full) is a large-scale synthetic dataset containing 3,500 designer-created houses with photorealistic rendered observations. The heavy presence of furniture, 3D object occlusions, and repetitive indoor layouts makes this dataset particularly challenging. ZInD is a real-world dataset comprising 1,575 unfurnished residential homes and 59,361 panoramic images. Following standard practice~\citep{grader2025supercharging}, panoramas are converted into perspective views for visual FLoc.

\renewcommand\arraystretch{0.8}
\begin{table*}[t] \small
\vspace{-0.3cm}
\caption{Comparative studies on the Gibson(f) and Gibson(g) datasets. * indicates the code is not open source and we reproduced the results by ourselves.}
\label{table1}
\centering
\setlength{\tabcolsep}{2.0mm}{
\begin{tabular}{l c c c c | c c c c }
\toprule
\multirow{2}{*}{\textbf{Method \tiny{(Venue)}}} &
\multicolumn{4}{c}{\textbf{Gibson(f) R@}} & \multicolumn{4}{c}{\textbf{Gibson(g) R@}} \\
\cline{2-9}
& \textbf{0.1 m}$\uparrow$  & \textbf{0.5 m}$\uparrow$  & \textbf{1 m}$\uparrow$ & \textbf{1 m 30$^\circ$}$\uparrow$ & \textbf{0.1 m}$\uparrow$ & \textbf{0.5 m}$\uparrow$ & \textbf{1 m}$\uparrow$  & \textbf{1 m 30$^\circ$}$\uparrow$ \\
\hline
PF-net\tiny{(CoRL 2018)} & 0 & 2.0 & 6.9 & 1.2 & 1.0 & 1.9 & 5.6 & 1.9 \\
MCL\tiny{(ICRA 1999)} & 1.6 & 4.9 & 12.1 & 8.2 & 2.3 & 6.2 & 9.7 & 7.3 \\
LASER\tiny{(CVPR 2022)} & 0.4 & 6.7 & 13.0 & 10.4 & 0.7 & 7.0 & 11.8 & 9.5 \\
F$^{3}$Loc\tiny{(CVPR 2024)} & 4.7 & 28.6 & 36.6 & 35.1 & 4.3 & 26.7 & 33.7 & 32.3 \\
3DP\tiny{(ACM MM 2025)} & 5.3 & 33.2 & 39.8 & 38.4 & 9.4 & 37.4 & 43.1 & 41.5 \\
RSK\tiny{(AAAI 2026)} & 8.3 & 38.5 & 45.3 & 43.6 & 8.7 & 36.4 & 42.3 & 40.4 \\
3DP \& RSK & 10.9 & 42.7 & 47.9 & 46.5 & 10.7 & 38.8 & 44.4 & 42.8 \\
UnLoc$^*$\tiny{(ICLR 2026)} & 11.8 & 45.0 & 49.6 & 48.3 & \textbf{12.3} & 45.0 & 49.9 & 48.2 \\
\hline
Ours & \textbf{11.96} & \textbf{49.48} & \textbf{57.25} & \textbf{56.00} & 11.88 & \textbf{48.23} & \textbf{55.11} & \textbf{53.51} \\

\bottomrule
\end{tabular}}
\vspace{-0.5cm}
\end{table*}

\textbf{Baselines.} We compare our method against four categories of baselines. 3DP and RSK retain their method-specific pretrained encoders, and our UnLoc baseline uses a Depth Anything V2 encoder with a DINOv2-S backbone. \emph{The implementation details of our method can be found in Appendix \ref{A1}.}

\begin{wraptable}{R}{0.59\textwidth} \small
\vspace{-15pt}
\caption{Comparative studies on the S3D (full) dataset. Sem. indicates whether semantics are used.}
\label{table2}
\centering
\setlength{\tabcolsep}{0.8mm}{
\begin{tabular}{l | c c c c | c}
\toprule
\multirow{2}{*}{\textbf{Method \tiny{(Venue)}}} &
\multicolumn{4}{c}{\textbf{S3D (full) R@}} & \multirow{2}{*}{\textbf{Sem.}} \\
\cline{2-5}
& \textbf{0.1 m}$\uparrow$  & \textbf{0.5 m}$\uparrow$  & \textbf{1 m}$\uparrow$  & \textbf{1 m 30$^\circ$}$\uparrow$ & \\
\hline
PF-net\tiny{(CoRL 2018)} & 0.2 & 1.3 & 3.2 & 0.9 & \multirow{9}{*}{no}  \\
MCL\tiny{(ICRA 1999)} & 1.3 & 5.2 & 7.8 & 6.4 & \\
LASER\tiny{(CVPR 2022)} & 0.7 & 6.4 & 10.4 & 8.7 &  \\
F$^{3}$Loc\tiny{(CVPR 2024)} & 1.5 & 14.6 & 22.4 & 21.3 &  \\
3DP\tiny{(ACM MM 2025)} & 5.6 & 27.4 & 55.5 & 24.0 &  \\
RSK\tiny{(AAAI 2026)} & 6.4 & 28.6 & 56.9 & 25.2 &  \\
3DP \& RSK & 6.7 &  26.8 & 54.7 & 24.2 & \\
UnLoc\tiny{(ICLR 2026)} & 5.5  & 34.2 & 40.4 & 39.3 \\
SceneAligner\tiny{(Arxiv 2026)} & 3.5  & 37.5 & 53.8 & 51.6 & \\
\hline
SemRayLoc$_s$\tiny{(ICCV 2025)} & 5.4 & 41.9 & 53.5 & 52.6 & \multirow{7}{*}{yes} \\
\  + 3DP\tiny{(ACM MM 2025)} & 5.5 & 46.6 & 56.7 & 56.2 &  \\
\  + RSK\tiny{(AAAI 2026)} & 6.2 & 48.1 & 59.9 & 58.8 &  \\
\  + 3DP \& RSK & 7.1 & 48.9 & 61.5 & 60.0 &  \\
SemRayLoc$_r$\tiny{(ICCV 2025)} & 5.7 & 45.5 & 58.8 & 57.5 &  \\
FoD$_d$\tiny{(CVPR 2026)} & 11.4 & 56.4 & 61.4 & 60.8 \\
FoD\tiny{(CVPR 2026)} & \textbf{12.0} & 65.2 & 71.9 & 71.4 \\
\hline
Ours (InfoNCE)	 & 8.9 & 52.6 & 60.8 & 59.9 & \multirow{2}{*}{no} \\
Ours (DisCo) & 10.0 & 64.4 & 75.0 & 74.3 & \\
\hline
SemRayLoc+DisCo & 9.0 & \textbf{67.5} & \textbf{81.5} & \textbf{81.1} & yes\\
\bottomrule
\end{tabular}}
\vspace{-0.5cm}
\end{wraptable}

\textbf{(1) Early Methods.} PF-net \citep{karkus2018particle}, MCL \citep{dellaert1999monte}, and LASER \citep{min2022laser}. \textbf{(2) Strong Baselines under the F$^{3}$Loc Framework.} F$^{3}$Loc \citep{chen2024f3loc} is a representative probabilistic method that combines a ray-based observation module with histogram filtering. 3DP \citep{chen20253dp} injects 3D geometric priors into the F$^{3}$Loc framework to improve localization accuracy without semantic labels. RSK \citep{chen2025perspective} employs unsupervised room-style knowledge learning to mitigate FLoc ambiguity. 3DP \& RSK adaptively leverages both geometric priors and room-style knowledge (\emph{see Appendix \ref{A2} for details}). UnLoc \citep{wuest2025unloc} employs a deep estimation encoder trained on large-scale data as the observation model and incorporates localization uncertainty. \textbf{(3) 3D Reconstruction-based Strong Baseline.} SceneAligner \citep{cho2026scenealigner} projects 3D reconstructions onto dense maps and aligns them with floorplans as a reconstruction-based visual FLoc baseline. \textbf{(4) Strong Baselines under the SemRayLoc Framework.} SemRayLoc$_s$ \citep{grader2025supercharging} leverages sparse floorplan semantics to predict semantic rays to reduce geometric ambiguity. By adapting visual pre-trainings from 3DP and RSK to SemRayLoc$_s$, three enhanced variants are derived: + 3DP, + RSK, and + 3DP \& RSK (\emph{see Appendix \ref{A3} for details}). SemRayLoc$_r$ further uses room-type labels; FoD \citep{ye2026fusion} fuses depth and floorplan semantics.

\textbf{Metrics.} Following existing work \citep{chen2024f3loc}, we report recall metrics computed at positional localization accuracies of 0.1 m, 0.5 m, and 1 m. To evaluate both positional and angular localization accuracy, we also report recall under a joint threshold of 1 m positional error and 30$^\circ$ orientation error. Recall is calculated as the percentage of predictions falling within these thresholds. 

\vspace{-0.3cm}
\subsection{Comparisons with SOTA Methods} \label{4.2}

\textbf{Results on Gibson(f) and Gibson(g).}
Tab.~\ref{table1} presents comparisons with SOTA methods on the Gibson benchmarks. Existing ray matching-based methods improve visual FLoc by generating geometry-consistent pose candidates, but remain vulnerable to structural aliasing among these candidates. DisCo consistently achieves the best performance at R@1 m and R@1 m 30$^\circ$ on both datasets. Specifically, DisCo improves R@1 m/R@1 m 30$^\circ$ to $57.25\%/56.00\%$ on Gibson(f) and $55.11\%/53.51\%$ on Gibson(g), outperforming the previous best semantic-free method UnLoc by large margins. This improvement demonstrates the effectiveness of our $SE(2)$-aware hypothesis preparation and contrastive disambiguation, which further distinguish ambiguous FLoc hypotheses beyond geometric matching.

\textbf{Results on S3D (full).}
Tab.~\ref{table2} reports results on the challenging S3D (full) dataset, which contains severe occlusions and repetitive indoor structures. Existing methods, including semantic-enhanced approaches, remain limited by ambiguous FLoc hypotheses generated from similar visual-floorplan patterns. DisCo achieves $75.0\%$ R@1 m and $74.3\%$ R@1 m30$^\circ$ without using any explicit semantic annotations, exceeding the listed baselines at these two thresholds. Compared with SceneAligner, DisCo improves R@1 m 30$^\circ$ by 22.7 percentage points, demonstrating the effectiveness of hypothesis preparation and $SE(2)$-aware contrastive disambiguation. Moreover, combining DisCo with SemRayLoc further boosts performance to $81.5\%$ R@1 m and $81.1\%$ R@1 m 30$^\circ$, validating that DisCo provides a general ambiguity resolution module complementary to different candidate generation strategies.

\begin{wraptable}{r}{0.55\textwidth} \small
\vspace{-11pt}
\caption{Comparative experimental results on the ZInD dataset. Sem. indicates whether to predict semantic rays during inference.}
\label{tab:zind}
\centering
\setlength{\tabcolsep}{2pt}
\begin{tabular}{lccccc}
\toprule
\textbf{Method\tiny{(Venue)}} & \textbf{Sem.} & \textbf{0.1 m}$\uparrow$ & 0.5 m$\uparrow$ & \textbf{1 m}$\uparrow$ & \textbf{1 m 30$^\circ$}$\uparrow$ \\
\midrule
LASER\tiny{(CVPR 2022)}     & No & 1.4          & 11.1          & 17.6          & 13.6          \\
F$^3$Loc\tiny{(CVPR 2024)}  & No & 0.7          & 7.9           & 15.1          & 11.5          \\
UnLoc$^*$\tiny{(ICLR 2026)}          & No &      1.5     &     11.4      &     16.7      &     13.8      \\
Ours       & No & 4.2 & 37.3 & 53.8 & 48.0 \\
\midrule
SemRayLoc\tiny{(ICCV 2025)} & Yes & 3.3           & 23.4          & 31.9          & 29.2          \\
SemRayLoc+DisCo        & Yes & 5.0 & 42.7 & 60.0 & 54.0 \\
\bottomrule
\end{tabular}
\vspace{-0.3cm}
\end{wraptable}

\textbf{Results on ZInD.}
Tab.~\ref{tab:zind} presents results on the real-world ZInD dataset. Despite the large appearance variations and complex residential layouts, DisCo achieves $53.8\%$ R@1 m and $48.0\%$ R@1 m 30$^\circ$ without using semantic annotations, exceeding the semantic-free baselines listed in the table at these thresholds. This demonstrates that the learned visual-floorplan compatibility can effectively capture implicit structural cues beyond explicit semantics. Furthermore, integrating DisCo with SemRayLoc improves R@0.5 m from $23.4\%$ to $42.7\%$ and improves R@1 m 30$^\circ$ from $29.2\%$ to $54.0\%$, validating its effectiveness as a general ambiguity resolution module for visual FLoc.

\begin{table}[h]
\vspace{-0.3cm}
\caption{
Ablation studies on S3D (full). 
(a) evaluates inner-floorplan position-level negatives (I-Pos-N) and orientation-level negatives (Ori-N), while retaining slack positives and cross-floorplan negatives. 
(b) evaluates slack positive sampling with positional perturbation (P-Pert) and angular perturbation (A-Pert), while retaining all negatives.
}
\label{tab:ablation}
\centering
\small
\setlength{\tabcolsep}{1.0mm}

\begin{minipage}{0.48\textwidth} \small
\centering
\textbf{(a) Within-floorplan hard negatives}

\begin{tabular}{cc|cccc} 
\toprule
\multicolumn{2}{c|}{\textbf{Ablations}} &
\multicolumn{4}{c}{\textbf{S3D (full) R@}} \\
\hline
I-Pos-N & Ori-N 
& \textbf{0.1 m}$\uparrow$ 
& \textbf{0.5 m}$\uparrow$ 
& \textbf{1 m}$\uparrow$ 
& \textbf{1 m 30$^\circ$}$\uparrow$ \\
\hline
\ding{55} & \ding{55} & 9.0 & 60.4 & 70.7 & 70.0 \\
\checkmark & \ding{55} & 9.5 & 62.4 & 73.5 & 72.8 \\
\ding{55} & \checkmark & 9.5 & 62.8 & 73.7 & 73.0 \\
\checkmark & \checkmark & 10.0 & 64.4 & 75.0 & 74.3 \\
\bottomrule
\end{tabular}
\end{minipage}
\hfill
\begin{minipage}{0.48\textwidth}
\centering
\textbf{(b) Slack positives}

\setlength{\tabcolsep}{1.0mm}
\begin{tabular}{cc|ccccc}
\toprule
\multicolumn{2}{c|}{\textbf{Ablations}} &
\multicolumn{4}{c}{\textbf{S3D (full) R@}} \\
\hline
P-Pert & A-Pert
& \textbf{0.1 m}$\uparrow$ 
& \textbf{0.5 m}$\uparrow$ 
& \textbf{1 m}$\uparrow$ 
& \textbf{1 m 30$^\circ$}$\uparrow$ \\
\hline
\checkmark & \ding{55} & 8.6 & 54.5 & 63.4 & 62.7 \\
\ding{55} & \checkmark & 8.9 & 57.8 & 67.0 & 66.1 \\
\checkmark & \checkmark & 10.0 & 64.4 & 75.0 & 74.3 \\
\bottomrule
\end{tabular}
\end{minipage}
\vspace{-0.3cm}
\end{table}

\subsection{Ablation Studies} \label{5.3}

\textbf{Effectiveness of Pose-Structured Contrastive Disambiguation.}
The InfoNCE baseline uses exact-pose positives and in-batch negatives. The variant without position-level and orientation-level negatives in Table~\ref{tab:ablation}(a) retains slack positives while using the same in-batch negatives. Compared with standard InfoNCE, DisCo improves R@1m from 60.8\% to 75.0\% and R@1 m 30$^\circ$ from 59.9\% to 74.3\% under the same candidate generation pipeline (Tab.~\ref{table2}). The significant gain verifies that uniform contrastive learning cannot effectively handle $SE(2)$-structured ambiguities. By introducing pose-aware positives and negatives, DisCo learns finer visual-floorplan compatibility and improves spatial and angular disambiguation.

\textbf{Effect of $SE(2)$-Aware Contrastive Constraints.} Tab.~\ref{tab:ablation}(a) evaluates the impact of within-floorplan hard negatives. Removing both position-level and orientation-level negatives degrades R@1 m 30$^\circ$ from $74.3\%$ to $70.0\%$, indicating that generic cross-floorplan negatives are insufficient for resolving structural aliasing. Introducing either position-level or orientation-level negatives consistently improves localization accuracy, while combining both achieves the best performance. This demonstrates that spatial and angular constraints provide complementary benefits for enhancing spatial separability and angular discriminability.

\textbf{Effect of Slack Positive Sampling.}
Tab.~\ref{tab:ablation}(b) investigates the influence of slack positives with different $SE(2)$ perturbations. Removing either positional or angular perturbations leads to noticeable performance drops, especially in the joint metric. Combining both perturbations improves R@1 m 30$^\circ$ from $62.7\%/66.1\%$ to $74.3\%$, demonstrating that slack positives effectively establish local pose neighborhoods and encourage smooth visual-floorplan compatibility under minor pose variations.

\begin{figure*}[!t]
 \centering
 \includegraphics[width=1.0\linewidth]{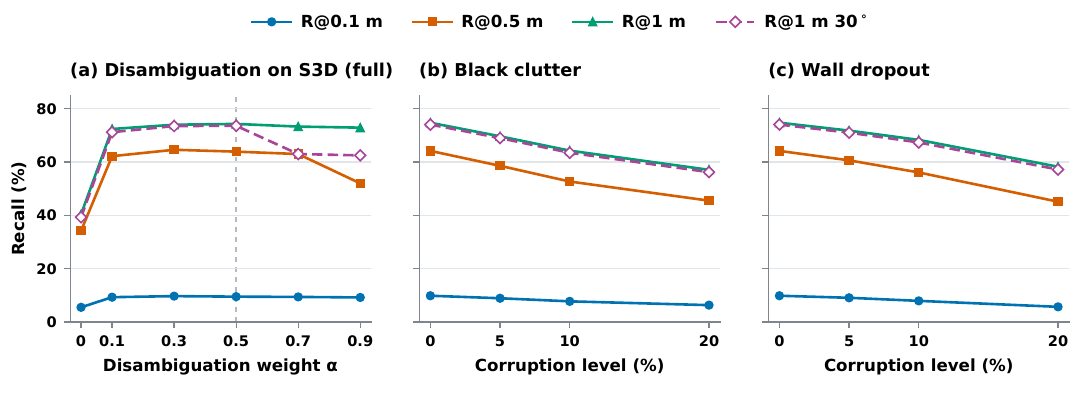}
 \vspace{-0.7cm}
 \caption{(a) An illustration of the parametric study of the disambiguation weight $\alpha$ on the S3D (full) dataset. (b) and (c) show that DisCo exhibits graceful degradation without crashing as the degree of corruption in the floorplan increases.}
 \label{alpha_robust}
 \vspace{-0.3cm}
\end{figure*}

\begin{table}[h] \small
\vspace{-0.3cm}
  \centering
  \caption{Experimental results for average hypothesis numbers, localization performance, and inference time when Top-$K$ takes different values.}
  \label{topk}
  \setlength{\tabcolsep}{1.0mm}
  \begin{tabular}{cccccc}
    \toprule
    \textbf{Top-$K$} & \textbf{Avg. Hypo.} & \textbf{Oracle R@1 m}$\uparrow$ & \textbf{R@1 m}$\uparrow$ & \textbf{R@1 m 30$^\circ$}$\uparrow$ & \textbf{Inference Time (ms)}$\downarrow$ \\
    \midrule
    50   & 8.7 & 83.20 & 67.90 & 66.98 & 26.16 $\pm$ 4.49 \\
    100  & 12.2 & 87.81 & 70.43 & 69.54 & 27.00 $\pm$ 5.48 \\
    250  & 20.2 & 92.29 & 72.65 & 71.82 & 29.95 $\pm$ 5.77 \\
    500  & 31.8 & 95.19 & 73.74 & 72.91 & 35.74 $\pm$ 6.88 \\
    1000 & 52.9 & 97.17 & 75.00 & 74.30 & 45.61 $\pm$ 8.32 \\
    \bottomrule
  \end{tabular}
  \vspace{-0.7cm}
\end{table}

\vspace{-0.3cm}
\subsection{Parametric Studies} \label{5.4}

\textbf{Impact of Top-$K$.} All ablation studies are conducted on the S3D (full) dataset. Tab.~\ref{topk} evaluates the impact of Top-$K$ on candidate coverage and efficiency. Larger Top-$K$ improves oracle recall from 83.20\% to 97.17\%, providing better coverage of the correct pose mode. After $SE(2)$-aware clustering, the retained FLoc hypotheses increase moderately, resulting in improved R@ 1m from 67.90\% to 75.00\%. Meanwhile, the inference time increases predictably from 26.16 ms to 45.61 ms, demonstrating that DisCo offers a controllable accuracy-efficiency trade-off and remains suitable for real-time localization. \emph{More detailed parametric studies and analyses can be found in Appendix \ref{C2}.} 

\textbf{Impact of Disambiguation Weight $\alpha$ and Local Floorplan Size.} As shown in Fig.~\ref{alpha_robust}(a), setting $\alpha$ to 0.3 or 0.5 helps achieve optimal visual FLoc performance. In this paper, we set $\alpha$ = 0.5. \emph{For parametric studies and analyses of the local floorplan size, please see Appendix \ref{C3}.} In this paper, we set the local floorplan size to 5 m $\times$ 5 m.

\textbf{Robustness to Imperfect Floorplans.}
We further evaluate the robustness of DisCo under imperfect floorplans by introducing two types of perturbations: black clutter, which adds spurious structures, and wall dropout, which removes partial wall structures. As shown in Tab.~\ref{alpha_robust}(b) and (c), DisCo exhibits graceful degradation as the corruption level increases. Even under severe 20\% corruption, the performance remains stable without collapse, achieving over 57\% R@1 m for both perturbation types. These results demonstrate that DisCo is robust to floorplan inaccuracies and environment-floorplan mismatches, benefiting from learned visual-floorplan compatibility rather than relying on exact geometric correspondence.

\emph{More computational efficiency analyses, comparative studies, visualizations, and discussion of failure cases can be found in Appendices \ref{D}, \ref{E}, \ref{F}, and \ref{G}, respectively.}

\vspace{-0.3cm}
\section{Conclusion}
In this work, we proposed DisCo, a general $SE(2)$-aware visual-floorplan contrastive disambiguation framework for visual FLoc. Different from existing methods that mainly focus on generating geometry-consistent pose candidates, DisCo explicitly addresses the structural ambiguity among candidate poses through clustering-based hypothesis preparation and pose-structured contrastive learning. The proposed framework improves spatial separability and angular discriminability without requiring explicit semantic annotations or scene-specific designs. Experiments across three benchmarks support the effectiveness of DisCo's pose-structured reranking, including substantial gains under fixed candidate-generation pipelines and strong 1 m and joint position--orientation recall on S3D (full). Moreover, DisCo exhibits strong robustness against imperfect floorplans, gracefully degrading rather than collapsing under substantial structural perturbations. We believe this work provides a general and reliable solution for visual FLoc and encourages future research on robust cross-modal scene understanding.

\section*{AI Use Statement}
We used generative AI tools to assist with building the experimental code framework, running experiments, and polishing the manuscript's language. The authors take responsibility for the experimental results, scientific claims, and final manuscript.


\bibliography{iclr2027_conference}
\bibliographystyle{iclr2027_conference}

\newpage
\appendix

\setcounter{tocdepth}{2} 

\begin{center}
{\large \textbf{Appendix Contents}}
\end{center}

\tableofcontents

\section{More Experimental Details}

\subsection{Implementation Details of Our Method} \label{A1}
To optimize the dense-to-ray geometric projection (RRP), we employ the Adam optimizer \citep{kingma2014adam} with a constant learning rate of $10^{-4}$ and a batch size of 64. The depth-aware visual encoder (DINOv2-S from Depth Anything V2 \citep{yang2024depth}) is frozen during training to preserve its depth estimation expertise. The remaining components are trained for 50 epochs on an NVIDIA RTX 3090 GPU. Our RRP matches 40 predicted rays to the floorplan to generate the initial geometry-aware pose likelihood map. For ray-depth discretization and candidate ray matching, we follow the same settings as F$^3$Loc \citep{chen2024f3loc}, including the values of $D$, $d_{\text{min}}$, $d_{\text{max}}$, $\gamma$, and the $SE(2)$ pose discretization. 

\textbf{Scoring and fusion details.} In Eq.~\ref{eq5}, the candidate-conditioned tokens and pooling weights are
\begin{equation*}
\begin{aligned}
\mathbf{H}_{j,l}&=\operatorname{LN}\!\left(\mathbf{F}_j+\operatorname{CrossAttn}(\mathbf{F}_j,\mathbf{G}_l,\mathbf{G}_l)\right),\\
a_{j,l,t}&=\operatorname{softmax}_t(\mathbf{w}^{\top}\mathbf{H}_{j,l,t}+b),\qquad
\operatorname{AttnPool}(\mathbf{H}_{j,l})=\sum_t a_{j,l,t}\mathbf{H}_{j,l,t}.
\end{aligned}
\end{equation*}
Here, $t$ indexes image tokens and $\theta$ denotes the trainable compatibility parameters. The MLP outputs an unnormalized scalar, with higher scores indicating better matches. During training, $\tau^{-1}=\exp(\beta)$ is learned with $\beta$ initialized to $\log(1/0.07)$. At inference, Eq.~\ref{eq9} is equivalent to maximizing $P_d(R_m|\mathcal{I}_j,F)\exp(\alpha P_c(R_m))$. Cached image tokens are reused, while cross-attention, pooling, and MLP scoring are candidate-specific.

\textbf{Compatibility network.} The frozen DINOv2-S backbone is loaded from the Depth Anything V2 checkpoint \citep{yang2024depth}. Its patch features are resized to a $6\times40$ grid and projected to 128 channels. Learned 2D positional encodings, layer normalization, and one four-head self-attention Transformer encoder layer produce the image tokens. The floorplan encoder retains the ResNet-18 spatial feature map, projects it to 128 channels, and adds learned 2D positional encodings and layer normalization. Four-head cross-attention uses image queries and floorplan keys/values, followed by a residual connection and layer normalization. A learned linear token scorer and a softmax over image tokens provide the pooling weights in Eq.~\ref{eq5}. The pooled vector passes through a scoring head consisting of layer normalization, a $128\rightarrow128$ linear layer, GELU, and a $128\rightarrow1$ linear layer. The score has no sigmoid or cosine normalization. The backbone is frozen; token projections, positional encodings, attention layers, pooling, and the scoring head are trainable. Training evaluates all image--candidate pairs in chunks of 16 candidate maps and applies row-wise cross-entropy to the temperature-scaled score matrix.

On S3D, the contrastive disambiguation module is trained using AdamW with a batch size of 32 for 30 epochs on an NVIDIA RTX 3090 GPU. We use a cosine learning-rate schedule with an initial learning rate of $10^{-4}$ and a minimum learning rate of $10^{-5}$. We select the checkpoint with the lowest loss on the validation split. Unless specified otherwise, the crop size for local floorplan structures is 5 m $\times$ 5 m. Each crop is centered at the sampled or proposed position and rotated by the inverse of its heading before floorplan encoding, aligning all candidate headings to a canonical direction. The same pose-conditioned crop construction is used for training samples and inference hypotheses. Cross-floorplan negatives are taken from other mini-batch samples and retained in all ablations. To encourage \textit{pose smoothness} via slack positive sampling, we apply positional and orientational perturbations of 0.5 m and $\pm 0.26$ radians to the GT pose, respectively. To improve \textit{spatial separability}, inner-floorplan position-level negative samples are randomly drawn within a distance range of 1.5 m to 3.0 m from the GT pose. To improve \textit{angular discriminability}, orientation-level negative samples are generated by applying a rotation sampled from $[30^\circ, 180^\circ]$ to the GT orientation. During inference, we first extract the top $K = 1000$ candidate poses from the DAFPM to ensure high candidate coverage. We then apply greedy $SE(2)$-aware candidate refinement with $\sigma_t = 0.6$ m, $\sigma_\theta = 30^\circ$, $\lambda_\theta = 1.0$, and $\rho = 1.0$ before DisCo reranking. We set the compatibility fusion weight in Eq.~9 to $\alpha=0.5$. All reported inference times are measured on an NVIDIA RTX 5880 Ada Generation GPU.

\begin{figure*}[h]
 \centering
 \includegraphics[width=0.9\linewidth]{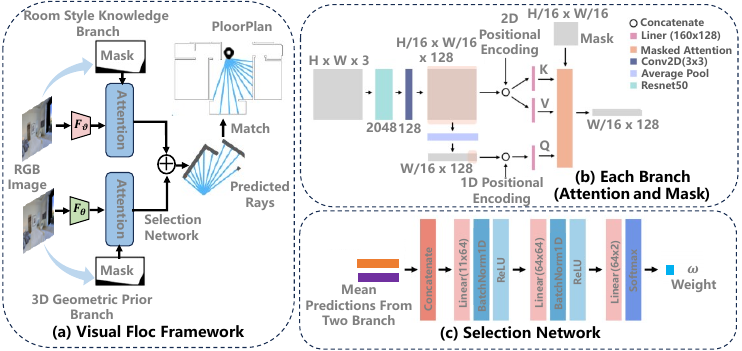}
 \caption{Illustrations of 3DP \& RSK. (a) The pre-trained visual encoders, $F_\theta$ in 3DP and $F_\vartheta$ in RSK, are transferred to the visual FLoc framework for fine-tuning to further fit this task. The visual FLoc framework is a dual-branch model consisting of a 3D geometric prior branch and a RSK branch. (b) and (c) detail the masked attention mechanism and the selection network, respectively.}
 \label{fig:supp_3dp_rsk}
 \vspace{-0.5cm}
\end{figure*}

\subsection{Implementation of 3DP \& RSK}
\label{A2}

3DP \& RSK is achieved by integrating the visual encoders pre-trained in 3DP \citep{chen20253dp} and RSK \citep{chen2025perspective} into a unified FLoc framework.
As shown in Fig.~\ref{fig:supp_3dp_rsk}(a), the fully pre-trained encoders, $F_\theta$ in 3DP and $F_\vartheta$ in RSK, are transferred to the visual FLoc framework for fine-tuning to fit the task, which localizes by finding the pose in the floorplan that has the most similar 2D rays as the prediction.
Our visual FLoc framework is implemented as a dual-branch model consisting of a 3D geometric prior branch and a RSK branch.
In each branch, the image is first aligned with the gravity direction, as done in \citep{chen2024f3loc}.
Then, $F_{\theta}$/$F_{\vartheta}$ and an attention \citep{vaswani2017attention}-based network are used to learn the probability distribution of planar depth over a range of depth hypotheses.
Pixels that become unobservable due to gravity alignment are masked in the attention, as shown in Fig.~\ref{fig:supp_3dp_rsk}(b).
To adaptively leverage 3D geometric priors and RSK based on the current view, a selection network, shown in Fig.~\ref{fig:supp_3dp_rsk}(c), implemented as a multilayer perceptron is adopted to learn a weight $0 \le \omega \le 1$ from the two predictions for adaptive selection:
\begin{equation}
\label{eq:supp_fusion}
\mathbf{P}_{\mathrm{Fusion}} = \omega \mathbf{P}_{\mathrm{3DP}} + (1 - \omega)\mathbf{P}_{\mathrm{RSK}}.
\end{equation}
$\mathbf{P}_{\mathrm{3DP}}$ and $\mathbf{P}_{\mathrm{RSK}}$ denote the probability distributions of planar depth from the 3D geometric prior branch and the RSK branch, respectively.
The expectation of $\mathbf{P}_{\mathrm{Fusion}}$ provides the final prediction of 2D rays.
$\omega$ is manually specified as 1 and 0, implying that only 3D geometric priors and RSK are used, respectively.
For the training of FLoc models, we optimize an L1 loss and a cosine-similarity-based shape loss:
\begin{equation}
\label{eq:supp_floc_loss}
\mathcal{L}_{\mathrm{FLoc}} =
\left\lVert \mathbf{d} - \mathbf{d}^{*} \right\rVert_1
+ 1 - \frac{\mathbf{d}^{\top}\mathbf{d}^{*}}
{\max\left\{\lVert \mathbf{d}\rVert_2 \lVert \mathbf{d}^{*}\rVert_2,\epsilon \right\}},
\end{equation}
where $\mathbf{d}$ and $\mathbf{d}^{*}$ are predicted and GT 2D-ray depths, respectively.
$\epsilon$ is a small constant to prevent division by zero.

\begin{figure*}[h]
 \centering
 \includegraphics[width=1.0\linewidth]{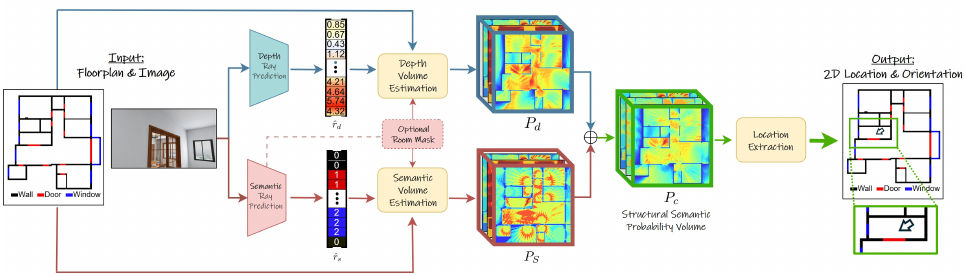}
 \caption{The pipeline of SemRayLoc from \citep{grader2025supercharging}.}
 \label{fig:supp_semrayloc}
 \vspace{-0.5cm}
\end{figure*}

\subsection{Variants of SemRayLoc}
\label{A3}

Similar to the implementation of 3DP \& RSK in Subsection~\ref{A2}, \textbf{SemRayLoc$_s$ + 3DP} is implemented by replacing the depth and semantic ray encoders in Fig.~\ref{fig:supp_semrayloc} with the fully pre-trained visual encoder $F_\theta$ from 3DP.
Similarly, \textbf{SemRayLoc$_s$ + RSK} is implemented by replacing the depth and semantic ray encoders in Fig.~\ref{fig:supp_semrayloc} with the fully pre-trained visual encoder $F_\vartheta$ from RSK.
\textbf{SemRayLoc$_s$ + 3DP \& RSK} is implemented by replacing the depth and semantic ray encoders with $F_\theta$ and $F_\vartheta$, respectively.

As shown in Tab.~\ref{tab:supp_weight}, increasing $\alpha$ from 0.0 substantially improves localization accuracy, confirming the importance of the learned visual-floorplan compatibility. The best overall performance is achieved at $\alpha=0.5$, while overly large weights degrade the joint positional-directional metric, suggesting that excessive reliance on the learned compatibility may weaken the stabilizing effect of the geometry-based likelihood.

\renewcommand\arraystretch{0.95}
\begin{table}[h] \small
  \centering
  \caption{A study on the effectiveness of DisCo under different datasets, candidate generation methods, and visual backbones.}
  \label{tab4}
  \setlength{\tabcolsep}{5pt}
  \begin{tabular}{lllccccc}
    \toprule
    \textbf{Dataset} & \textbf{Candidate} & \textbf{Backbone} & \textbf{DisCo} & \textbf{R@0.1 m}$\uparrow$ & \textbf{R@0.5 m}$\uparrow$ & \textbf{R@1 m}$\uparrow$ & \textbf{R@1 m 30$^\circ$}$\uparrow$ \\
    \midrule
    S3D (full) & RRP & DINOv2 & \ding{55} & 5.5 & 34.2 & 40.4 & 39.3 \\
    S3D (full) & RRP & DINOv2 & \checkmark & \textbf{10.0} & \textbf{64.4} & \textbf{75.0} & \textbf{74.3} \\
    \midrule
    S3D (full) & $\text{F}^3\text{Loc-style}$ & ResNet-50 & \ding{55} & 2.56 & 22.26 & 31.79 & 30.49 \\
    S3D (full) & $\text{F}^3\text{Loc-style}$ & ResNet-50 & \checkmark & \textbf{6.62} & \textbf{52.46} & \textbf{63.87} & \textbf{62.83} \\
    \midrule
    Gibson(f)    & $\text{F}^3\text{Loc-style}$       & ResNet-50 & \ding{55} & 4.63 & 28.55 & 36.62 & 35.03 \\
    Gibson(f)    & $\text{F}^3\text{Loc-style}$       & ResNet-50 & \checkmark & \textbf{7.17} & \textbf{40.06} & \textbf{48.38} & \textbf{46.94} \\
    \bottomrule
  \end{tabular}
  \vspace{-0.5cm}
\end{table}

\section{Additional Controlled Studies} \label{B}


To investigate whether the effectiveness of DisCo mainly comes from the backbone or the candidate generation module, we conduct a controlled study by integrating DisCo with different candidate generators and visual backbones. As shown in Tab.~\ref{tab4}, DisCo consistently improves localization performance across different settings, demonstrating its generality as a pose disambiguation module.

First, we compare RRP with and without DisCo under exactly the same candidate generator, image encoder, checkpoint, and candidate poses. The only difference is whether DisCo is applied for pose refinement. On S3D, DisCo improves R@1m from 40.4\% to 75.0\% and R@1m,30$^\circ$ from 39.3\% to 74.3\%, indicating that the performance gain mainly originates from the proposed SE(2)-aware disambiguation mechanism rather than a stronger visual backbone.

Furthermore, we integrate DisCo with an F$^3$Loc-style candidate generator using a ResNet-50 backbone. Without changing the upstream localization pipeline, DisCo improves R@1m from 31.79\% to 63.87\% on S3D and from 36.62\% to 48.38\% on Gibson(f). These results demonstrate that DisCo is not tied to a specific backbone or proposal strategy, but can consistently enhance different localization pipelines by resolving ambiguous pose hypotheses.

Overall, these experiments verify that candidate generation and pose disambiguation are complementary components. While the candidate generator provides high-recall pose hypotheses, DisCo effectively improves their spatial and angular separability through SE(2)-aware contrastive learning, serving as a general refinement module for visual floorplan localization.

\begin{table}[h] \small
  \centering
  \caption{Experimental results for the average hypothesis numbers and localization performance when the parameters in Eq.~\ref{eq4} are changed.}
  \label{tab:parameter_ablation}
  \setlength{\tabcolsep}{2.0mm}
  \begin{tabular}{crcc}
    \toprule
    \textbf{Setting} & \textbf{Avg. Hypotheses} & \textbf{R@1 m}$\uparrow$ & \textbf{R@1 m 30$^\circ$}$\uparrow$ \\
    \midrule
    $\sigma_t = 0.3\text{ m}$   & 121.3 & 75.22 & 74.55 \\
    $\sigma_t = 0.6\text{ m}$   &  52.9 & 75.00 & 74.30 \\
    $\sigma_t = 1.0\text{ m}$   &  33.3 & 74.32 & 73.49 \\
    \midrule
    $\sigma_\theta = 15^\circ$  &  62.7 & 74.91 & 74.22 \\
    $\sigma_\theta = 30^\circ$  &  52.9 & 75.00 & 74.30 \\
    $\sigma_\theta = 45^\circ$  &  47.8 & 74.75 & 74.00 \\
    \midrule
    $\lambda_\theta = 0.5$      &  48.0 & 74.71 & 73.96 \\
    $\lambda_\theta = 1.0$      &  52.9 & 75.00 & 74.30 \\
    $\lambda_\theta = 2.0$      &  57.3 & 74.71 & 73.99 \\
    \midrule
    $\rho = 0.5$               & 132.8 & 75.11 & 74.52 \\
    $\rho = 1.0$               &  52.9 & 75.00 & 74.30 \\
    $\rho = 1.5$               &  31.6 & 74.40 & 73.52 \\
    \bottomrule
  \end{tabular}
  \vspace{-0.5cm}
\end{table}

\section{Additional Parametric Studies}

\subsection{Study of the Parameters in Eq.~\ref{eq4}} \label{C1}

We investigate the sensitivity of the key parameters in Eq.~\ref{eq4} in the main paper, including $\sigma_t$, $\sigma_\theta$, $\lambda_\theta$, and $\rho$. As shown in Tab.~\ref{tab:parameter_ablation}, DisCo maintains stable performance under different parameter settings, demonstrating the robustness of the proposed SE(2)-aware refinement.

For $\sigma_t$ and $\rho$, smaller values retain more hypotheses and slightly improve localization accuracy, while larger values reduce the hypothesis number with only marginal performance degradation. For example, changing $\rho$ from 0.5 to 1.5 reduces the average hypotheses from 132.8 to 31.6, while R@1m only decreases from 75.11\% to 74.40\%. Similarly, varying $\sigma_\theta$ and $\lambda_\theta$ results in negligible performance differences, indicating that DisCo is insensitive to the exact angular refinement settings.

Overall, these results demonstrate that DisCo provides a flexible accuracy-efficiency trade-off by adjusting the retained hypothesis number, while maintaining robust localization performance across different refinement configurations.

\begin{table}[h] \small
  \centering
  \caption{Experimental results for average hypothesis numbers, localization performance, and inference time when Top-$K$ takes different values.}
  \label{tab:source_topk}
  \setlength{\tabcolsep}{1.0mm}
  \begin{tabular}{cccccc}
    \toprule
    \textbf{Top-$K$} & \textbf{Avg. Hypo.} & \textbf{Oracle R@1 m}$\uparrow$ & \textbf{R@1 m}$\uparrow$ & \textbf{R@1 m 30$^\circ$}$\uparrow$ & \textbf{Inference Time (ms)}$\downarrow$ \\
    \midrule
    50   & 8.7 & 83.20 & 67.90 & 66.98 & 26.16 $\pm$ 4.49 \\
    100  & 12.2 & 87.81 & 70.43 & 69.54 & 27.00 $\pm$ 5.48 \\
    250  & 20.2 & 92.29 & 72.65 & 71.82 & 29.95 $\pm$ 5.77 \\
    500  & 31.8 & 95.19 & 73.74 & 72.91 & 35.74 $\pm$ 6.88 \\
    1000 & 52.9 & \textbf{97.17} & \textbf{75.00} & \textbf{74.30} & 45.61 $\pm$ 8.32 \\
    \bottomrule
  \end{tabular}
  \vspace{-0.5cm}
\end{table}

\begin{table}[h] \small
  \centering
  \caption{Experimental results on the average hypothesis numbers, localization performance, and inference time when the maximum hypothesis number is fixed.}
  \label{tab:max_hypotheses}
  \setlength{\tabcolsep}{1.0mm}
  \begin{tabular}{ccccccc}
    \toprule
    \textbf{Max. Hypo.} & \textbf{Avg. Hypo.} & \textbf{R@0.1 m}$\uparrow$ & \textbf{R@0.5 m} & \textbf{R@1 m}$\uparrow$ & \textbf{R@1 m 30$^\circ$}$\uparrow$ & \textbf{Inference Time (ms)}$\downarrow$ \\
    \midrule
    16       & 16.0 & 9.51          & 61.05          & 71.18          & 70.34          & 31.35 $\pm$ 3.98 \\
    32       & 32.0 & 9.73          & 63.19          & 73.52          & 72.77          & 37.22 $\pm$ 4.05 \\
    48       & 45.4 & 9.80          & 64.03          & 74.52          & 73.72          & 42.40 $\pm$ 4.95 \\
    64       & 51.1 & 9.84 & 64.18 & 75.00 & 74.30          & 44.77 $\pm$ 6.03 \\
    Uncapped & 52.9 & \textbf{10.00} & \textbf{64.40} & \textbf{75.00}          & \textbf{74.30} & 46.62 $\pm$ 10.86 \\
    \bottomrule
  \end{tabular}
  \vspace{-0.5cm}
\end{table}

\subsection{Study on Top-$K$ and the Maximum Number of Hypotheses} \label{C2}

We further investigate the influence of the hypothesis budget on localization performance and efficiency. As shown in Tab.~\ref{tab:source_topk}, increasing the source Top-K mainly improves the proposal recall. Specifically, increasing Top-K from 50 to 1000 improves Oracle R@1m from 83.20\% to 97.17\% and final R@1m from 67.90\% to 75.00\%, at the cost of increased inference time. This demonstrates that the proposed framework provides a flexible accuracy-efficiency trade-off by adjusting the candidate budget.

We further analyze the maximum number of refined hypotheses in Tab.~\ref{tab:max_hypotheses}. The computational cost is mainly determined by the number of representative hypotheses after SE(2)-aware selection rather than the initial candidate pool. Reducing the hypothesis budget from 64 to 16 decreases the inference time from 44.77 ms to 31.35 ms, while maintaining competitive performance (R@1m: 71.18\% vs. 75.00\%). 

Overall, these results demonstrate that DisCo achieves controllable scalability: the candidate budget improves robustness against missing pose modes, while the refined hypothesis number provides a direct trade-off between localization accuracy and inference efficiency.

\begin{table}[h] \small
\centering
\caption{Additional parametric study on the disambiguation weight $\alpha$ on the S3D (full) dataset.}
\label{tab:supp_weight}
\setlength{\tabcolsep}{2.0mm}
\begin{tabular}{c|cccc}
\toprule
\textbf{Disam. Weight} & \textbf{R@0.1 m}$\uparrow$ & \textbf{R@0.5 m}$\uparrow$ & \textbf{R@1 m}$\uparrow$ & \textbf{R@1 m 30$^\circ$}$\uparrow$ \\
\midrule
0.0 & 5.5 & 34.2 & 40.4 & 39.3 \\
0.1 & 9.3 & 62.2 & 72.4 & 71.2 \\
0.3 & 9.7 & 64.6 & 74.0 & 73.5 \\
0.5 & 10.0 & 64.4 & 75.0 & 74.3 \\
0.7 & 9.4 & 63.0 & 73.3 & 63.0 \\
0.9 & 9.2 & 52.0 & 72.9 & 62.5 \\
\bottomrule
\end{tabular}
\vspace{-0.5cm}
\end{table}

\begin{table}[h] \small
\centering
\caption{Additional parameter study on the crop size of floorplan structures on the S3D (full) dataset.}
\label{tab:supp_crop}
\setlength{\tabcolsep}{2.0mm}
\begin{tabular}{c|cccc}
\toprule
\textbf{Crop Size} & \textbf{R@0.1 m}$\uparrow$ & \textbf{R@0.5 m}$\uparrow$ & \textbf{R@1 m}$\uparrow$ & \textbf{R@1 m 30$^\circ$}$\uparrow$ \\
\midrule
3 m $\times$ 3 m & 8.7 & 55.6 & 65.7 & 64.5 \\
5 m $\times$ 5 m & 10.0 & 64.4 & 75.0 & 74.3 \\
7 m $\times$ 7 m & 9.7 & 62.5 & 72.5 & 71.8 \\
\bottomrule
\end{tabular}
\vspace{-0.5cm}
\end{table}

\subsection{Study on the Disambiguation Weight and Local FloorPlan Size} \label{C3}

In this section, we provide additional parametric studies on the disambiguation weight $\alpha$ and the crop size of local floorplan structures. As discussed in the main paper, we evaluate these parameters on the S3D (full) dataset. As shown in Fig. \ref{tab:supp_weight}, setting $\alpha$ to 0.3 or 0.5 helps achieve optimal visual FLoc performance. Tab.~\ref{tab:supp_crop} shows that a 5 m $\times$ 5 m crop achieves the best overall performance. Smaller crops may miss sufficient surrounding layout context, whereas larger crops can introduce irrelevant structures from neighboring regions, making visual-floorplan matching less discriminative.

\begin{table}[h] \small
\centering
\small
\caption{Average single-query inference time (ms) on an NVIDIA RTX 5880 Ada Generation GPU. The default DisCo result is taken from the Top-$K=1000$ setting in Tab.~\ref{topk}.}
\label{tab:runtime}
\setlength{\tabcolsep}{6mm}
\begin{tabular}{l c}
\toprule
\textbf{Method} & \textbf{Time (ms)} $\downarrow$ \\
\hline
F$^3$Loc & 6.4 \\
SemRayLoc & 169.4 \\
RRP & 11.56 \\
Ours & 45.61 $\pm$ 8.32 \\
\bottomrule
\end{tabular}
\vspace{-0.5cm}
\end{table}

\section{Analysis of Computational Efficiency} \label{D}

We evaluate single-query inference time on an NVIDIA RTX 5880 Ada Generation GPU (Tab.~\ref{tab:runtime}). F$^3$Loc and RRP require 6.4 ms and 11.56 ms per query, respectively. With the default Top-$K=1000$ setting, DisCo retains an average of 52.9 hypotheses and takes $45.61\pm8.32$ ms per query, corresponding to approximately 21.9 queries per second. SemRayLoc takes approximately 3.7 times as long per query (169.4 ms). This configuration achieves 75.0\% R@1 m and 74.3\% R@1 m 30$^\circ$ on S3D (full). The hypothesis-budget study in Tab.~\ref{tab:max_hypotheses} further shows that capping the candidate set at 48 hypotheses reduces mean latency to 42.40 ms, with 74.52\% R@1 m, providing a controllable accuracy-efficiency trade-off.

\begin{table}[h] \small
\centering
\caption{Comparative studies of long-sequence trajectory tracking methods on the Gibson(t) dataset.}
\label{tab:supp_tracking}
\setlength{\tabcolsep}{2.0mm}
\begin{tabular}{lcccc}
\toprule
\multirow{2}{*}{\textbf{Method \tiny{(Venue)}}} &
\multicolumn{4}{c}{\textbf{Gibson(t) R@}} \\
\cmidrule(lr){2-5}
& \textbf{R@0.2 m}$\uparrow$ & \textbf{R@1 m}$\uparrow$ & \textbf{RMSE(S)}$\downarrow$ & \textbf{RMSE(A)}$\downarrow$ \\
\midrule
LASER\tiny{(CVPR 2022)} & - & 59.5 & 0.39 & 1.96 \\
F$^{3}$Loc\tiny{(CVPR 2024)} & 35.1 & 89.2 & 0.18 & 0.88 \\
F$^{3}$Loc fusion\tiny{(CVPR 2024)} & 62.2 & 94.6 & 0.12 & 0.51 \\
3DP\tiny{(ACM MM 2025)} & 70.3 & 97.3 & 0.12 & 0.34 \\
RSK\tiny{(AAAI 2026)} & 59.5 & 94.6 & 0.13 & 0.51 \\
3DP \& RSK & 64.9 & 94.6 & 0.12 & 0.51 \\
\midrule
Ours & 73.0 & 94.6 & 0.11 & 0.49 \\
\midrule
Oracle & - & 100.0 & 0.07 & 0.07 \\
\bottomrule
\end{tabular}
\vspace{-0.5cm}
\end{table}

\section{Additional Evaluation for Long-Sequence Trajectory Tracking} \label{E}

In this section, we compare our method with existing SOTA methods on the long-sequence trajectory tracking task using the Gibson(t) dataset collected by F$^3$Loc \citep{chen2024f3loc}.
Gibson(t) consists of 118 long image sequences, each containing 280 $\sim$ 5152 image frames.
The \textbf{R}oot-\textbf{M}ean-\textbf{S}quare \textbf{E}rror (RMSE) over the last 10 frames is employed to measure the accuracy of sequential trajectory tracking when localization is successful (RMSE(S)) and in all cases (RMSE(A)).
Technically, we combine the histogram filter proposed by F$^3$Loc with our method and perform visual FLoc with 100 historical frames.

As shown in Tab.~\ref{tab:supp_tracking}, our method improves recall by 2.7\% compared to 3DP at the localization accuracy of 0.2 m.
The reduction in the RMSE(S) metric reflects the robustness of our method in sequential trajectory tracking.
Additionally, our method achieves competitive results in terms of 1 m localization accuracy and RMSE(A).
It is worth noting that F$^3$Loc fusion performs visual FLoc by adaptively utilizing single-frame and multi-frame images, yet it is only competitive on the RMSE(S) metric.
Oracle achieved 100\% Recall at 1 m accuracy by combining GT rays with a histogram filter, demonstrating the potential of visual FLoc.

\section{More Visualizations} \label{F}

\begin{figure*}[t]
    \centering
    \setlength{\tabcolsep}{1pt}
    \renewcommand{\arraystretch}{0.9}
    \newcommand{\imgw}{0.17\linewidth}

    \begin{tabular}{c c c c c c}

        \small RGB Input
        & \includegraphics[width=\imgw]{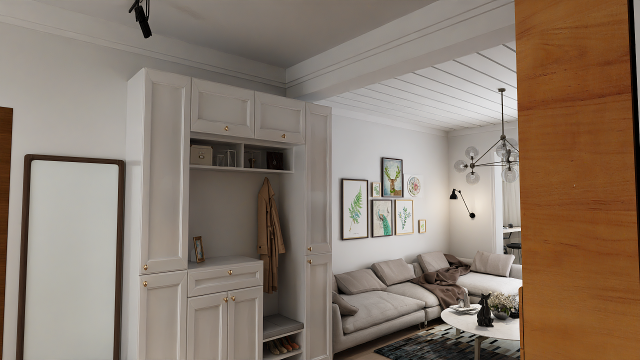}
        & \includegraphics[width=\imgw]{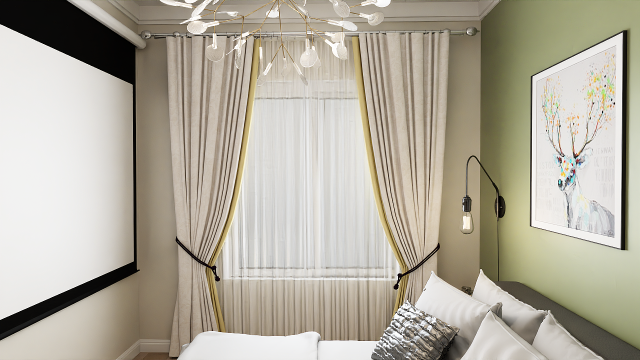}
        & \includegraphics[width=\imgw]{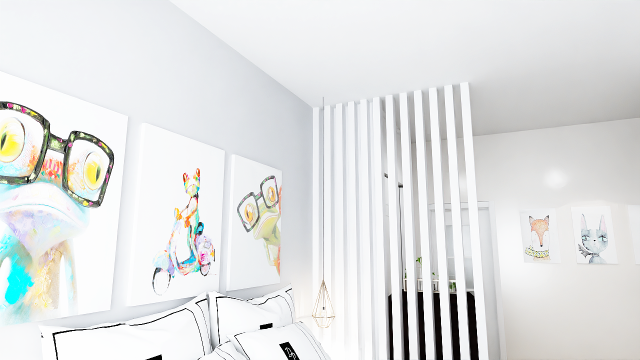}
        & \includegraphics[width=\imgw]{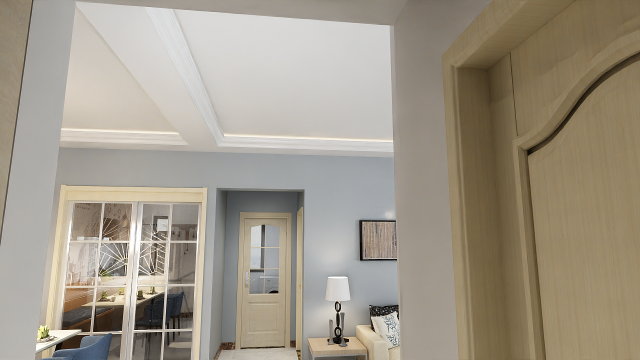}
        & \includegraphics[width=\imgw]{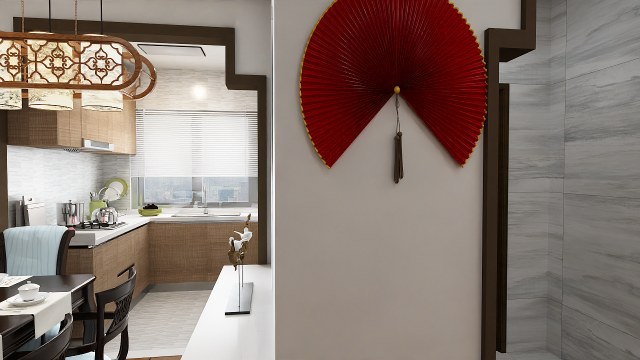} \\

        \small SemRayLoc
        & \includegraphics[width=\imgw]{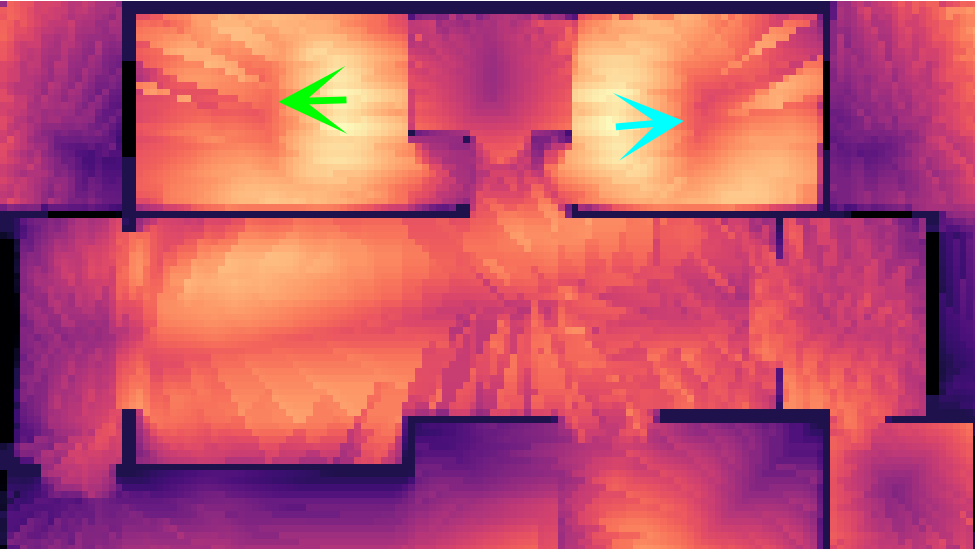}
        & \includegraphics[width=\imgw]{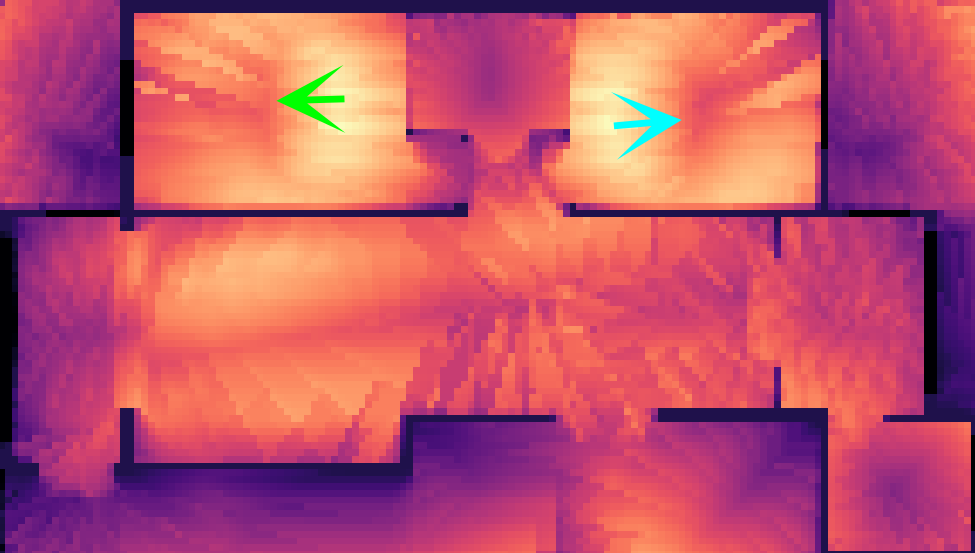}
        & \includegraphics[width=\imgw]{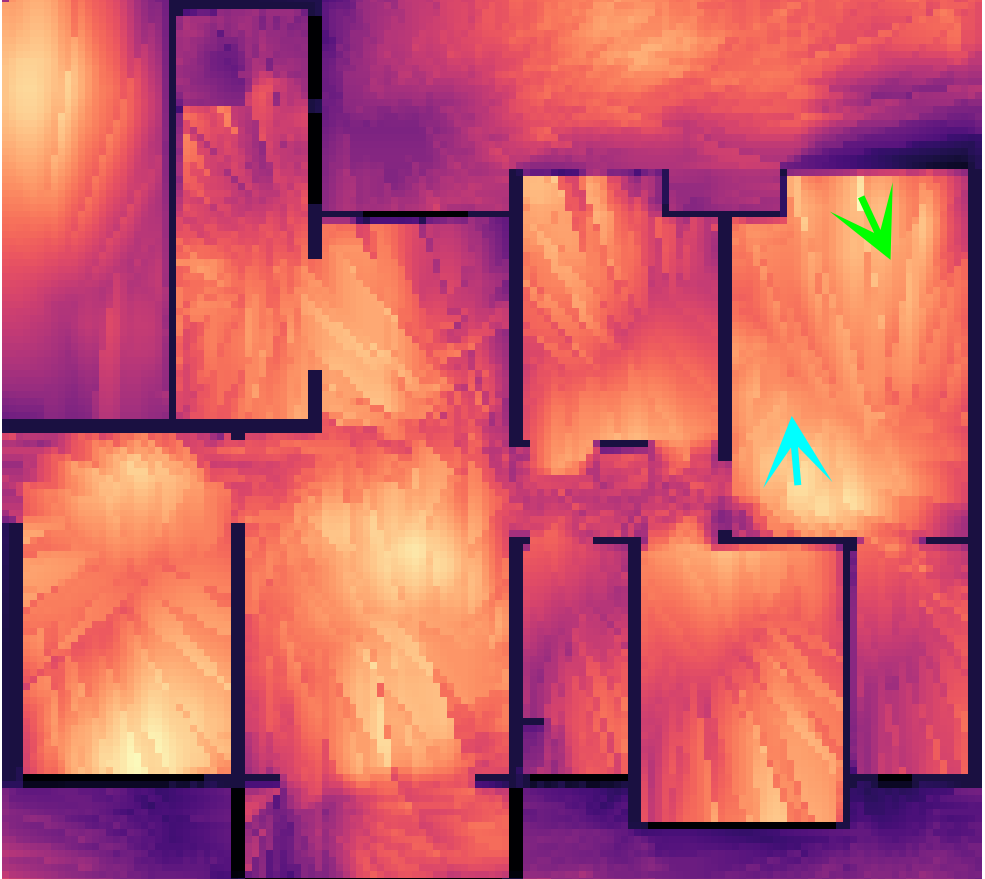}
        & \includegraphics[width=\imgw]{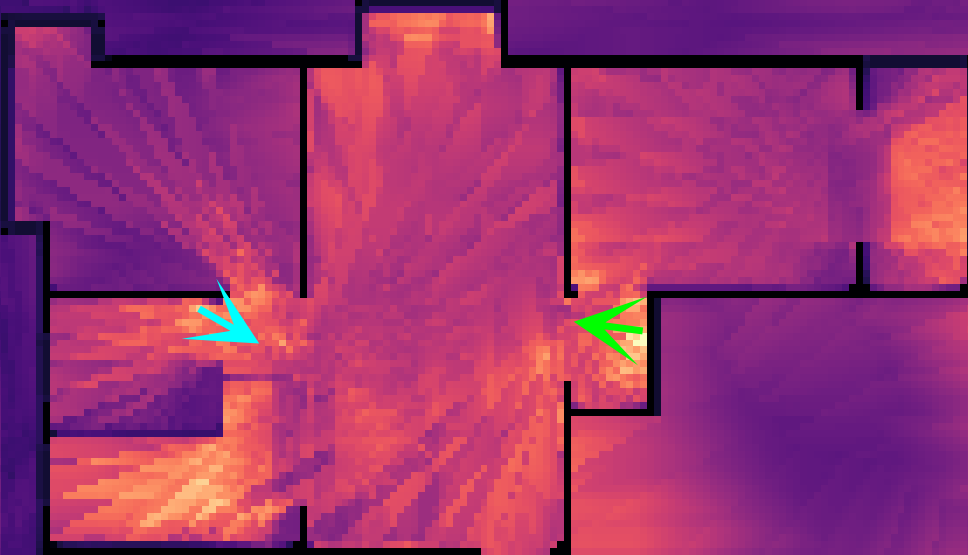}
        & \includegraphics[width=\imgw]{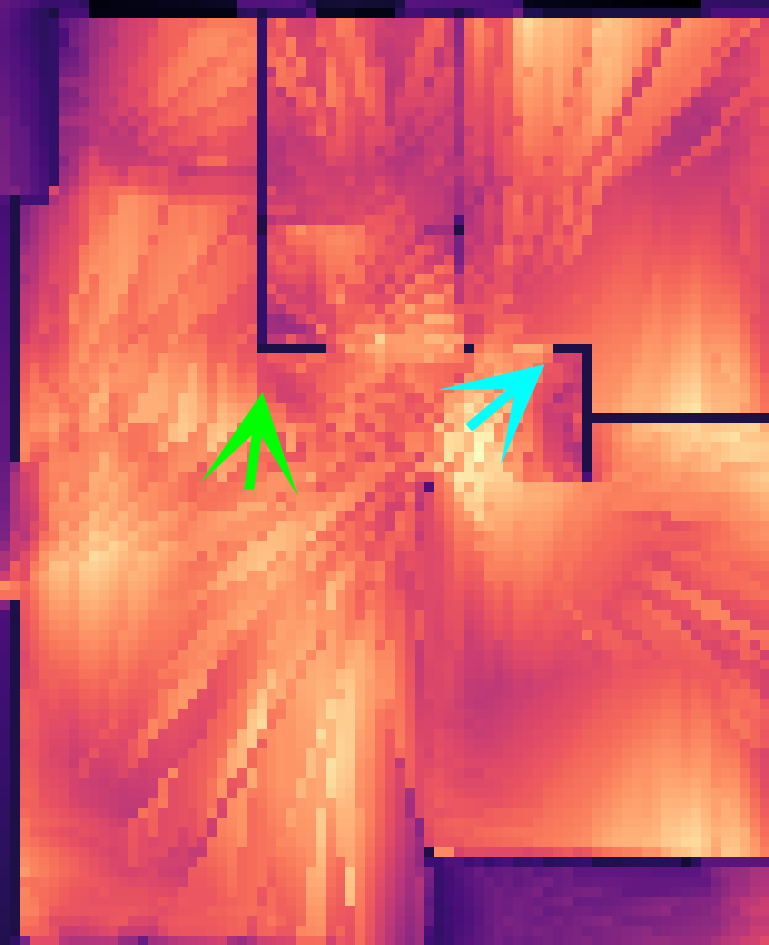} \\

        \small Our RRP
        & \includegraphics[width=\imgw]{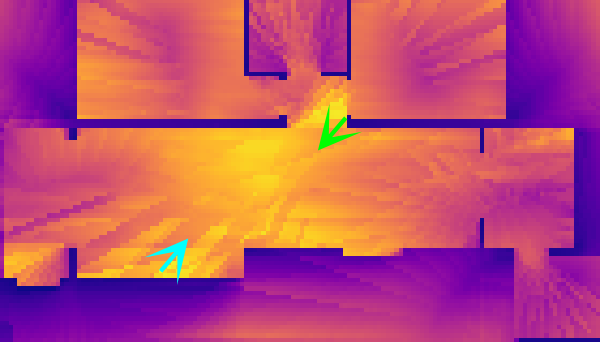}
        & \includegraphics[width=\imgw]{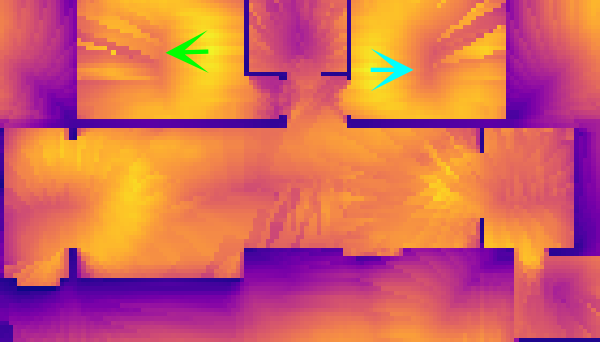}
        & \includegraphics[width=\imgw]{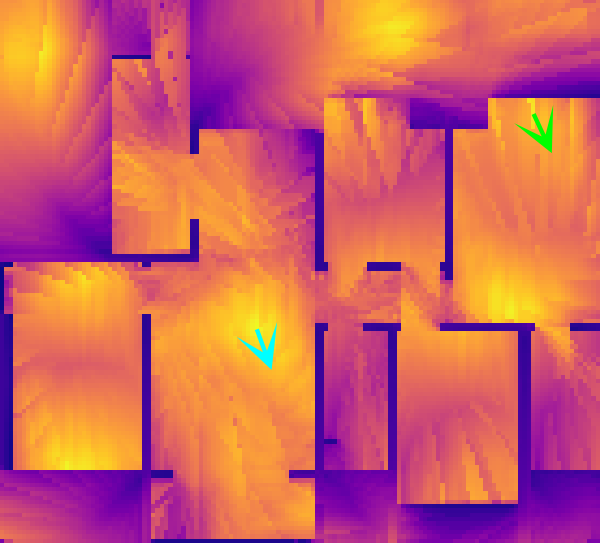}
        & \includegraphics[width=\imgw]{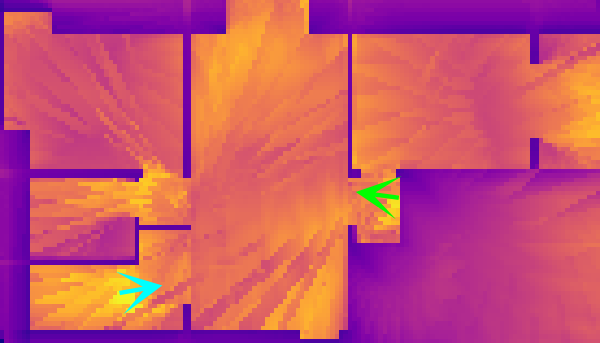}
        & \includegraphics[width=\imgw]{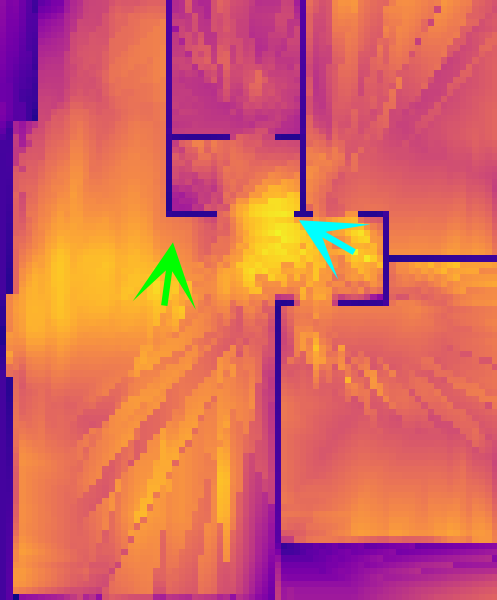} \\

        \small Our DPM
        & \includegraphics[width=\imgw]{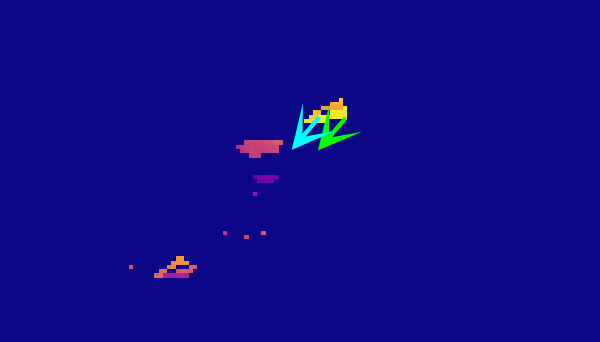}
        & \includegraphics[width=\imgw]{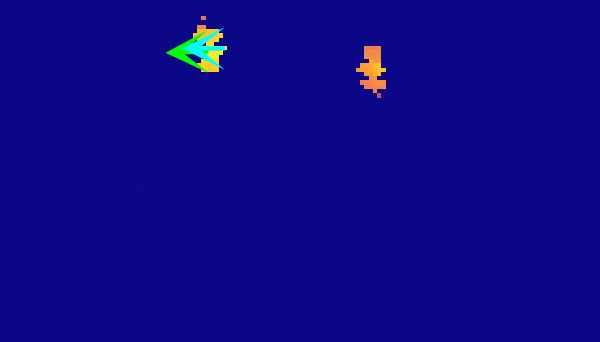}
        & \includegraphics[width=\imgw]{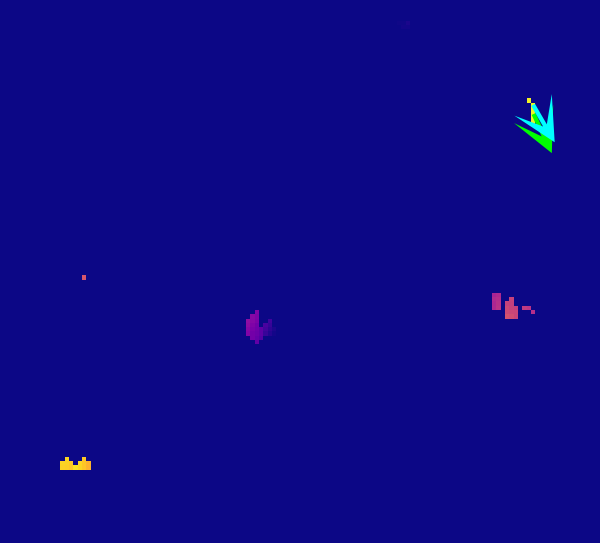}
        & \includegraphics[width=\imgw]{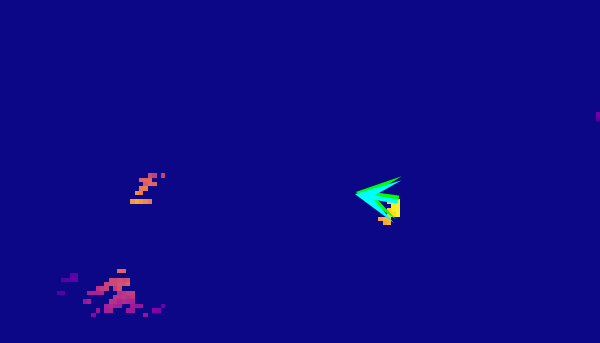}
        & \includegraphics[width=\imgw]{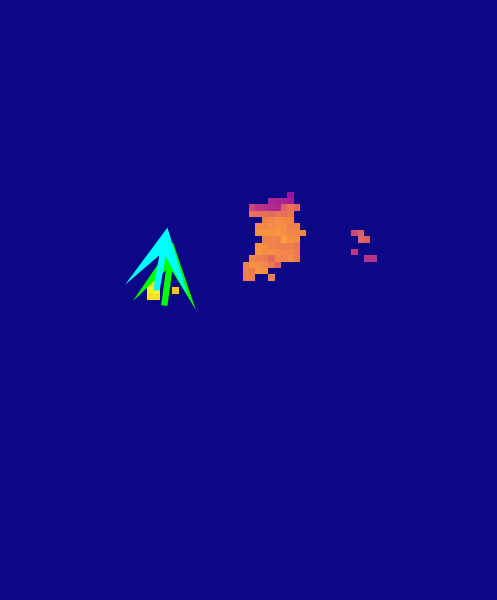} \\

        \small Final FLoc
        & \includegraphics[width=\imgw]{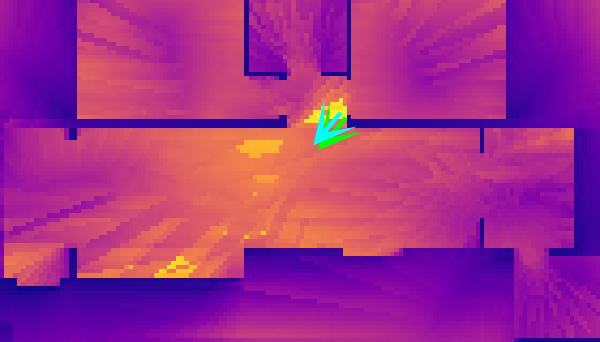}
        & \includegraphics[width=\imgw]{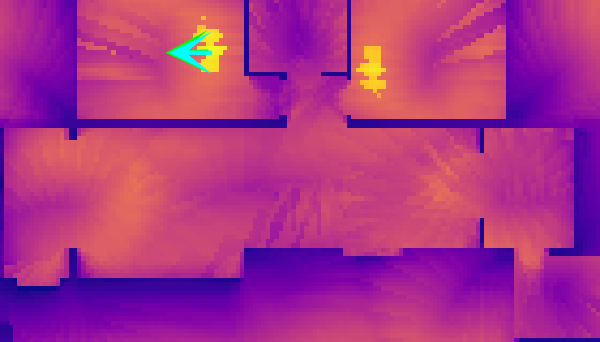}
        & \includegraphics[width=\imgw]{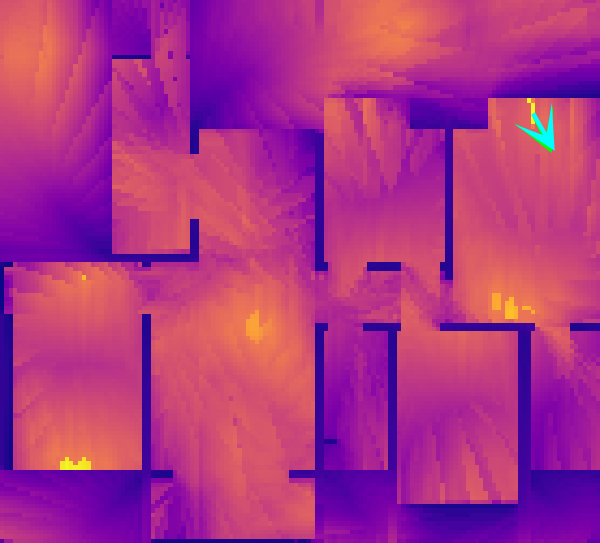}
        & \includegraphics[width=\imgw]{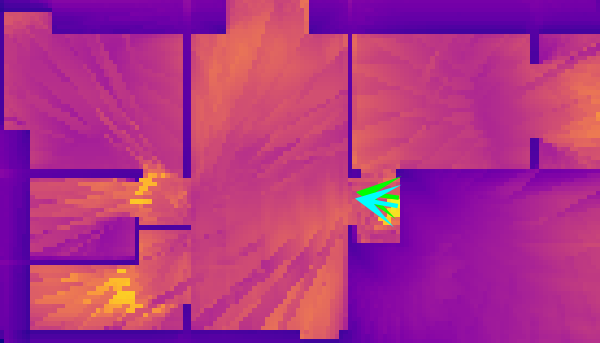}
        & \includegraphics[width=\imgw]{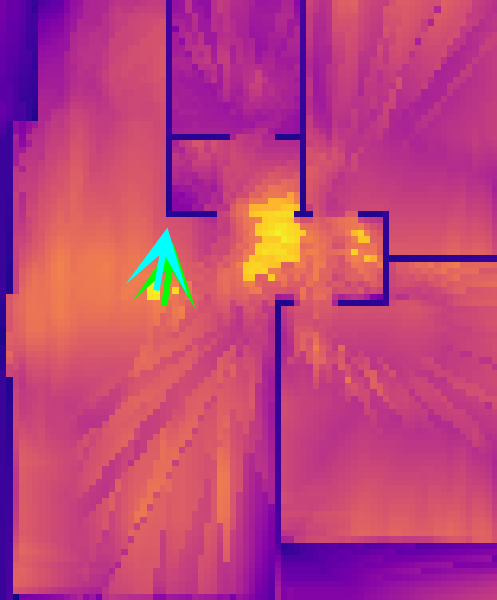} \\
    \end{tabular}

    \caption{Qualitative comparison on 5 scenes.}
    \label{fig:supp_visualization}
\end{figure*}

Fig.~\ref{fig:supp_visualization} shows qualitative comparisons between our method and SemRayLoc \citep{grader2025supercharging}.
Due to the repetitive geometric structures in the floorplan, our RRP also fails when SemRayLoc encounters errors.
However, our visual-floorplan contrastive disambiguation effectively addresses localization ambiguities and yields highly accurate final FLoc results.

\section{Failure Cases} \label{G}

Although DisCo effectively resolves structural ambiguities in semantic-free visual FLoc, it may still fail in several challenging scenarios. For example, when different regions exhibit highly similar geometric layouts but distinct functional semantics (e.g., room categories or object identities), semantic-based methods such as SemRayLoc can leverage additional scene context and achieve better discrimination. DisCo may also struggle when the floorplan geometry provides insufficient discriminative information, such as extremely repetitive structures or significant appearance–geometry mismatches.

These cases highlight the complementary nature of semantic information and our proposed pose disambiguation mechanism. DisCo focuses on improving spatial and angular separability without requiring costly semantic annotations, making it applicable to semantic-free settings. Integrating semantic cues or high-level scene priors into the candidate generation or refinement stages provides a promising direction for further improving robustness in such challenging scenarios.

\end{document}